\documentclass[journal]{IEEEtran}
\usepackage[utf8]{inputenc} % allow utf-8 input
\usepackage[T1]{fontenc}    % use 8-bit T1 fonts
\usepackage{hyperref}       % hyperlinks
\usepackage{url}            % simple URL typesetting
\usepackage{booktabs}       % professional-quality tables
\usepackage{amsfonts}       % blackboard math symbols
\usepackage{nicefrac}       % compact symbols for 1/2, etc.
\usepackage{microtype}      % microtypography
\usepackage{graphicx}
\usepackage{subfigure}
\usepackage{booktabs}
\usepackage{dashrule}
\usepackage{color}
\usepackage{colortbl}
\definecolor{mygray}{gray}{.9}

\usepackage{multicol}

\usepackage{graphicx}
\usepackage{comment}
\usepackage{amsmath,amssymb} % define this before the line numbering.
\usepackage{color}
\usepackage{mathrsfs}
\usepackage{url}
\usepackage{graphicx}
\usepackage{subfigure}
\usepackage{multirow}
\usepackage{bm}
\usepackage{amssymb}
\usepackage{animate}
\usepackage{multicol}
\usepackage{algorithmic}
\usepackage{algorithm}
\usepackage{amsmath}
\usepackage{caption}
\usepackage[numbers,sort&compress]{natbib}

\usepackage{hyperref}
\hypersetup{
	colorlinks=true,
	linkcolor=blue,
	filecolor=blue,      
	urlcolor=blue,
	citecolor=cyan,
}

\ifCLASSINFOpdf
\else
\fi
\begin{document}
%\title{Task-agnostic Temporally Consistent \\ Facial Video Editing}
\title{UniFaceGAN: A Unified Framework for Temporally Consistent Facial Video Editing}

\author{Meng~Cao\footnotemark*,
        Haozhi Huang\footnotemark*,
        Hao Wang,
        Xuan Wang,
        Li Shen,
        Sheng Wang, %~\IEEEmembership{Member,~IEEE,}
        Linchao Bao\footnotemark$\dagger$,
        Zhifeng Li,~\IEEEmembership{Senior Member,~IEEE,}
        Jiebo Luo,~\IEEEmembership{Fellow,~IEEE}% <-this % stops a space
%\thanks{This work is done when Meng Cao works as an intern at Tencent AI Lab.}% <-this % stops a space
\thanks{Meng Cao is with Peking University, Beijing, China. Haozhi Huang and Sheng Wang are with Xverse, Shenzhen, China. Li Shen is with JD Explore Academy, Beijing, China. Hao Wang, Xuan Wang, Sheng Wang, Linchao Bao and Zhifeng Li are with Tencent AI Lab, Shenzhen, China. Jiebo Luo is with Department of Computer Science, University of Rochester, Rochester, NY 14627, USA.}
}


\markboth{Journal of \LaTeX\ Class Files,~Vol.~14, No.~8, August~2015}%
{Shell \MakeLowercase{\textit{et al.}}: Bare Demo of IEEEtran.cls for IEEE Journals}



\maketitle
\renewcommand{\thefootnote}{\fnsymbol{footnote}} 
\footnotetext[1]{These authors contributed equally to this work.}
\footnotetext[2]{Corresponding Author.}
\begin{abstract}
Recent research has witnessed advances in facial image editing tasks including face swapping and face reenactment. However, these methods are confined to dealing with one specific task at a time. In addition, for video facial editing, previous methods either simply apply transformations frame by frame or utilize multiple frames in a concatenated or iterative fashion, which leads to noticeable visual flickers. In this paper, we propose a unified temporally consistent facial video editing framework termed UniFaceGAN. Based on a 3D reconstruction model and a simple yet efficient dynamic training sample selection mechanism, our framework is designed to handle face swapping and face reenactment simultaneously. To enforce the temporal consistency, a novel 3D temporal loss constraint is introduced based on the barycentric coordinate interpolation. Besides, we propose a region-aware conditional normalization layer to replace the traditional AdaIN or SPADE to synthesize more context-harmonious results. Compared with the state-of-the-art facial image editing methods, our framework generates video portraits that are more photo-realistic and temporally smooth. Our project is available at \href{https://sites.google.com/view/mengcao/publication/unifacegan}{here}.
\end{abstract}


% \fi
\IEEEpeerreviewmaketitle

\section{Introduction}
\begin{figure*}[t]
	\centering
	\includegraphics[width=0.75\linewidth]{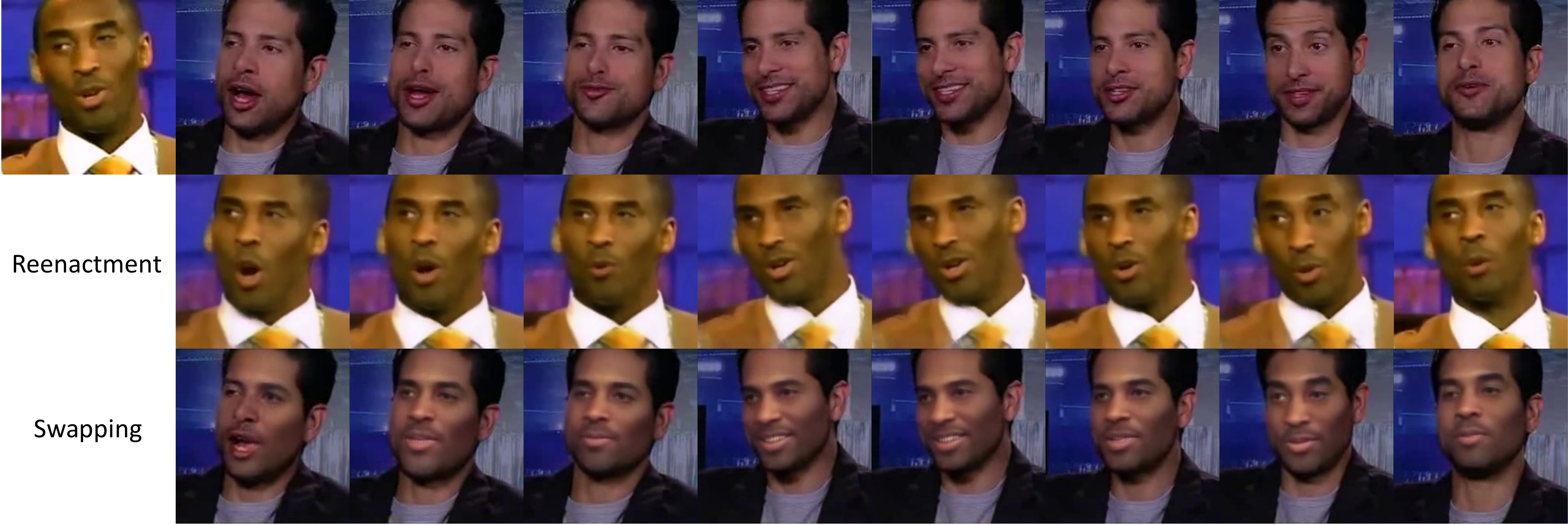}
    \caption{\small{Results of UniFaceGAN. Face Reenactment and Swapping.}}
    \vspace{-3mm}
	\label{fig:teaser}
\end{figure*}
\IEEEPARstart{F}{acial} video editing is an extensively studied task in the industry due to its widespread applications in movie composition, computer games, and privacy protection. It includes face swapping~\cite{kim2018deep,olszewski2017realistic,korshunova2017fast,nirkin2018face,jin2017cyclegan,nirkin2019fsgan}, face reenactment~\cite{garrido2014automatic,thies2016face2face,suwajanakorn2017synthesizing,averbuch2017bringing,wiles2018x2face,siarohin2019animating,wu2018reenactgan,zakharov2019few,zhao2017dual}, $etc$. Face swapping aims at replacing one person's identity with another while keeping the original attributes unchanged, \textit{e.g.} head pose, facial expression, background, \textit{etc}. Face reenactment is another task that focuses on transferring the facial poses and expressions from the others.

Most of the previous works have regarded face swapping and face reenactment as separate tasks and could only handle either of them. In this work, we devote ourselves to developing a more unified framework termed UniFaceGAN which settles face swapping and face reenactment jointly. Besides accomplishing the existing tasks, more surprisingly, our method also supports a novel application, which disentangles and remixes the facial characteristics from three different individuals. Specifically, we recombine the identity, expression, and appearance information extracted from three individual image frames to synthesize a photo-realistic video portrait. Notably, we name it ``appearance" as it contains both the background and the pose information. We refer to this novel editing application as ``fully disentangled manipulation" in our paper.

Despite tremendous advances in image facial editing, it is still challenging to perform video-level editing because of the temporal diversity. Simply applying transformations frame by frame inevitably induces temporal inconsistency, such as visual flickers. To address this issue, several methods have been developed. For face swapping, FSGAN~\cite{nirkin2019fsgan} generates the output in an iterative way. However, reusing the previous frames recursively incurs error accumulation, which brings about more distortion. For face reenactment, \cite{zakharov2019few,wiles2018x2face} regularize temporal relationships implicitly by concatenating multiple frames before feeding them to the networks. However, these methods lack explicit supervision. In this paper, we introduce a novel 3D temporal loss based on the dense optical flow map via the barycentric coordinate interpolation, which provides explicit supervision to guarantee the temporally consistent outputs. 

UniFaceGAN has three cascaded procedures, namely Dynamic Training Sample Selection, 3D Disentangled Editing, and Deep Blending Network. The Dynamic Training Sample Selection mechanism is designed to sample training data in both intra-video and inter-video manners, which facilitates our multi-task training. In the 3D Disentangled Editing stage, we incorporate a 3D Morphable Model (3DMM)~\cite{blanz1999morphable,zhu2016face} reconstruction module to decompose a video portrait into pose, expression, and identity coefficients. Then we recombine these factors according to the specific task to get a transformed face sequence through a rasterization renderer. As for the Deep Blending Network, we feed the rendered transformed face and the auxiliary appearance hint to a Generative Adversarial Network, which translates the input to the photo-realistic result. 

We're not the first to introduce the 3D reconstruction technology into facial editing and the pioneering works~\cite{kim2018deep,nagano2018pagan} also adopt a 3D-to-2D generation idea for accomplishing face reenactment for a single pre-determined person. However, our UniFaceGAN differs in the following two respects. Firstly, UniFaceGAN takes an additional appearance image as input and unifies the solutions of multiple video portrait manipulation tasks for arbitrary persons in a multi-task manner. Besides, the appearance image also serves as a background hint in the 3D Disentangled Editing stage, which settles the problem of blur background and generates more photo-realistic outputs.
%with a dynamic training sample selection mechanism. Specifically, two facial video editing tasks are jointly trained in an adversarial way with a mixture of image and video datasets. While image datasets have more varieties in identities, video datasets can provide temporal constraints. %A dynamic training sample pick-up strategy is adopted to facilitate the hybrid training process. 
Introducing the \textbf{3D Disentangled Editing} stage brings several advantages. First, compared to 2D-based methods~\cite{zakharov2019few,ha2019marionette,nirkin2019fsgan,li2019faceshifter}, explicitly reconstructing a 3D face can naturally collect that information in the 3D space. Besides, 3DMM has already decomposed pose, expression and identity information, which can be recombined to generate a novel desired 3D face. On the contrary, the pose and expression information in 2D facial landmarks cannot be decoupled, leading to the limited manipulation freedom. In addition, it is worth mentioning that sparse 2D landmarks lack facial texture details, thus suffering from blurry generation results. In contrast, rendering the 3DMM model with texture details provides more informative and realistic hints.% for the latter GAN-Based Portrait Generation stage. 

Since the facial information comes from the rendered face and the appearance image provides the background hint, \textbf{Deep Blending Network} uses two separate networks to tackle these two kinds of clues. Specifically, it consists of an encoder-decoder backbone and an appearance embedder, serving as the facial and non-facial feature extractors respectively. The encoder-decoder backbone takes a rendered 3D face as input, while the appearance embedder maps the concatenation of the appearance image and the non-facial area binary mask into a low-resolution feature map. Then, the facial and non-facial clues are fused through our proposed region-aware conditional normalization layer (RCN) which precisely controls the feature-level mixture of the face and background.

%Extensive experimental results show that UniFaceGAN generates impressive outputs with high fidelity and particularly achieves the desired temporal consistency on multiple video portrait manipulation tasks. 
The contributions of this work are summarized as follows:

\begin{itemize}
	%\item With 3D Disentangled Editing and Dynamic Training Sample Selection Mechanism, we present a unified framework that offers solutions for multiple tasks, including face swapping, face reenactment, and `Fully disentangled manipulation', in the same network. Experiment results show this multi-task learning mechanism is efficient and improve the performance on both.
	\item We present a unified framework that offers solutions for multiple tasks, including face swapping, face reenactment, and ``fully disentangled manipulation". A simple yet efficient Dynamic Training Sample Selection Mechanism is adopted to aid this multi-task learning procedure.
	%\item We conduct the network training in a hybrid manner, which fully utilizes the variety of image datasets and the temporal consistency of video datasets.
	\item  To achieve better video coherence, a novel 3D optical flow loss is introduced based on the barycentric coordinate interpolation.
	
	%we offer a novel scheme for optical flow map estimation  and introduce a temporal loss for.
	\item  A region-aware conditional normalization (RCN) layer tailored for our task is proposed which modulates the features in the decoding process in a spatial-aware way and results in more authentic outputs.
\end{itemize}

\begin{figure*}[t]
	\centering
	\includegraphics[width=0.8\linewidth]{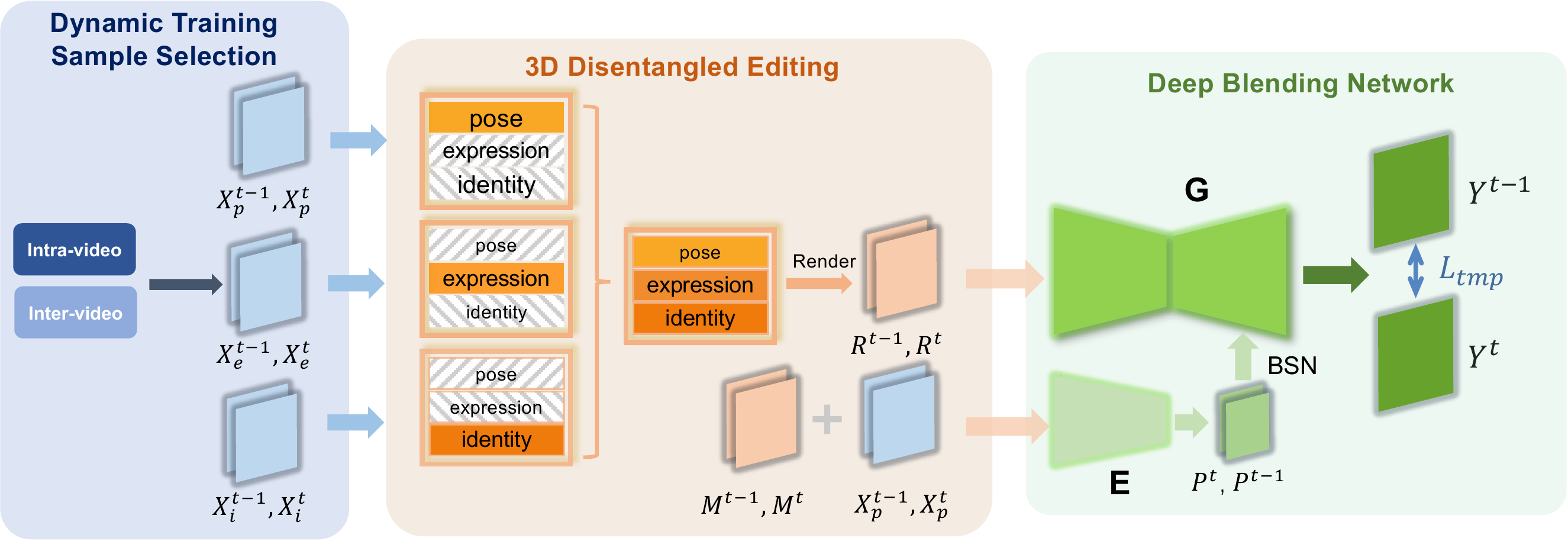}
	\caption{\small{The pipeline of our proposed UniFaceGAN.}}
	\vspace{-3mm}
	\label{fig:pipeline}
\end{figure*}

\section{Related Work}
%In Section~\ref{subsec:editing}, we briefly review the recent remarkable achievements of face swapping and face reenactment including both 3D reconstruction-based and 2D landmark-based methods. Then in Section~\ref{subsec:temporalConsitency}, we analyze the approaches current facial editing methods use to enforce the temporal consistency. %Finally, Section~\ref{subsec:cnorm} describes some typical conditional normalization layers.
\subsection{Facial Image Editing}\label{subsec:editing}
\noindent\textbf{Face Swapping.}
Methods for face swapping were proposed as far back as 2004 by~\cite{blanz2004exchanging}, which considers 3D transformations between two faces but requires manual involvement. \cite{olszewski2017realistic} and \cite{nirkin2018face} firstly reconstruct 3D faces for inputs and then conduct image blending.
%Methods for face swapping were proposed as far back as 2004 by~\cite{blanz2004exchanging}, which considers 3D transformations between two faces but requires manual involvement. Olszewski et al.~\cite{olszewski2017realistic} fit a multi-linear PCA model to obtain the 3D geometry and perform texture and shape retargeting to achieve video face swapping. Nirkin et al.~\cite{nirkin2018face} achieve face swapping results for images by reconstructing 3D faces for inputs and conducting image blending. However,~\cite{nirkin2018face} requires the laborious pixel-level annotation labels and an additional segmentation network to adjust the facial boundary while our UniFaceGAN utilizes the GAN-based blending network to achieve the boundary fusion with no additional data needed.
For 2D-based methods, Korshunova et al.~\cite{korshunova2017fast} formulate the face swapping as a 2D style transfer problem. Jin et al.~\cite{jin2017cyclegan} utilize CycleGAN to transfer facial expressions and head poses more consistently. IPGAN~\cite{bao2018towards} uses two separate encoder networks to achieve the identity and attribute embedder vectors and a generator based on the concatenation of them. %These methods, however, either use embedding space or latent vectors to represent the identity and attribute information, which lacks explicit guidance and is only driven by various loss functions. In contrast, our UniFacaGAN uses 3D Morphable Model (3DMM)~\cite{blanz1999morphable,zhu2016face} which is pre-trained with the large-scale AFLW2000-3D dataset~\cite{zhu2017face} to decomposes the facial image in a more disentangling and explicit way. 

\noindent\textbf{Face Reenactment.} 
Face2Face~\cite{thies2016face2face} fits 3DMM to both source and target faces and then transfers the expressions from one to another by exchanging corresponding coefficients. %Suwajanakorn et al.~\cite{suwajanakorn2017synthesizing} synthesize the mouth shape directly from audio relying on a recurrent neural network. 
PAGAN~\cite{nagano2018pagan} uses the subject’s expression blendshapes and linearly blends these key textures to achieve fine-scale manipulation. However, the output of PAGAN is limited in the pure black background, which hinders its practicality. \cite{kim2018deep} adopts a 3D head representation and a novel space-time neural network to achieve photo-realistic face reenactment for a pre-defined person. %However, the space-time neural network of~\cite{kim2018deep} has no background hint as input, which makes it confined to one specific task and suffers from the blurred background. Conversely, the Deep Blending Network of UniFaceGAN has the appearance image as input and therefore generates more realistic outputs.
As for the 2D-based method, ~\cite{garrido2014automatic} retrieves the image based on the temporal clustering of target frames and a novel matching metric. %Averbuch-Elor et al.~\cite{averbuch2017bringing} uses 2D warping and generates fine-scale dynamic details to animate target images.
X2Face~\cite{wiles2018x2face} is exploited to learn an embedded face representation and to map from this embedded face representation to the generated frame. Monkey-Net~\cite{siarohin2019animating} aims to encode motion information via moving keypoints learned in a self-supervised fashion. %Nonetheless, these warping-based methods often fail and have noticeable artifacts when dealing with large-scale pose movements.
ReenactGAN~\cite{wu2018reenactgan} maps the source face onto a boundary latent space and transforms the source face’s boundary to the target's one.  
%Zakharov et al.~\cite{zakharov2019few} propose a few-shot neural talking head model to generate high-fidelity outputs only using a few or one image after the lengthy meta-learning on a large video dataset. MarioNETte~\cite{ha2019marionette} focuses on the identity preserving problem which frequently accrues in the face reenactment and proposes a landmark transform technique to produce reenactments of unseen identities. %\cite{zakharov2020fast} decomposing a person's appearance through a pose-dependent coarse image and a pose-independent texture image to generate more photo-realistic results. \cite{geng2020towards} also adopts a similar 3D decomposition and recombination strategy. However, \cite{geng2020towards} focuses on the facial expression editing task which can only imitate expressions, not poses.

%We propose to address face swapping and face reenactment in a unified framework, which is rarely discussed in previous publications. To our best knowledge, only FSGAN implements this feature. However, FSGAN tends to generate outputs with blurry background. Benefitting from the 3D reconstruction technique, our UniFaceGAN improves the output quality greatly. Besides, we also introduce a novel application called ``fully disentangled manipulation" which takes the pose, identity and expression from three different images to synthesis photo-realistic results. Previous methods including FSGAN are unable to handle this case. % may delete for space

\subsection{Temporal Consistency in Facial Editing}\label{subsec:temporalConsitency}
Several publications investigate the temporal consistency problem. FSGAN~\cite{nirkin2019fsgan} used an iterative generation way in face reenactment and segmentation stages, namely previous generation results are used for the prediction of the next frames. \cite{zakharov2019few} and \cite{ha2019marionette} propose a few-shot neural talking head model based on the concatenation of several frames.

However,  iterative refinement tends to propagate the previous frame artifacts to the next one whereas the simple concatenation manner lacks explicit supervision. In contrast, our proposed UniFaceGAN explicitly uses a temporal loss to regularize consecutive frames. Besides, a hybrid training strategy with our proposed Dynamic Training Sample Selection mechanism is also adopted to implicitly enforce the  temporal consistency.

\section{Proposed Approach}
\label{headings}
The overall pipeline is shown in Fig.~\ref{fig:pipeline}. $X_i^t$, $X_p^t$ and $X_e^t$ denote the identity image, appearance image, and expression image at time step $t$, respectively. %Firstly, we elaborate on the Dynamic Training Sample Selection mechanism in Section~\ref{subsec:Select}. Then 3D Disentangled Editing is described in Section~\ref{subsec:3D}. Finally, we present the Deep Blending Network and the loss function design in Section~\ref{subsec:GAN} and Section~\ref{subsec:loss} respectively.

\subsection{Dynamic Training Sample Selection}\label{subsec:Select}

\begin{figure*}[t]
	\centering
	\subfigure[]{
		\begin{minipage}[b]{0.5\linewidth}
			\includegraphics[height=0.5\linewidth]{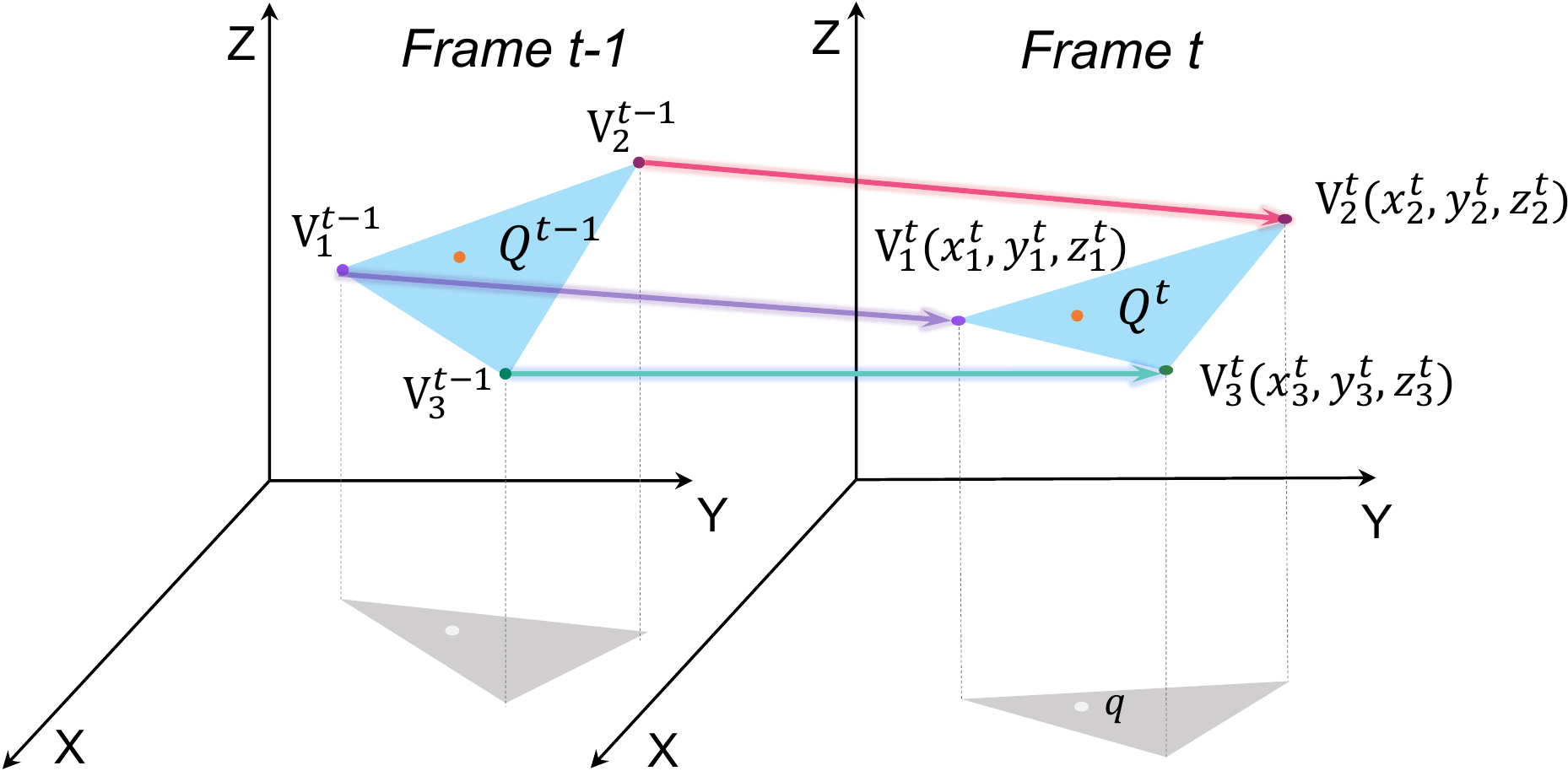}
		\end{minipage}
	}
	\subfigure[]{
		\begin{minipage}[b]{0.4\linewidth}
			\includegraphics[height=0.6\linewidth]{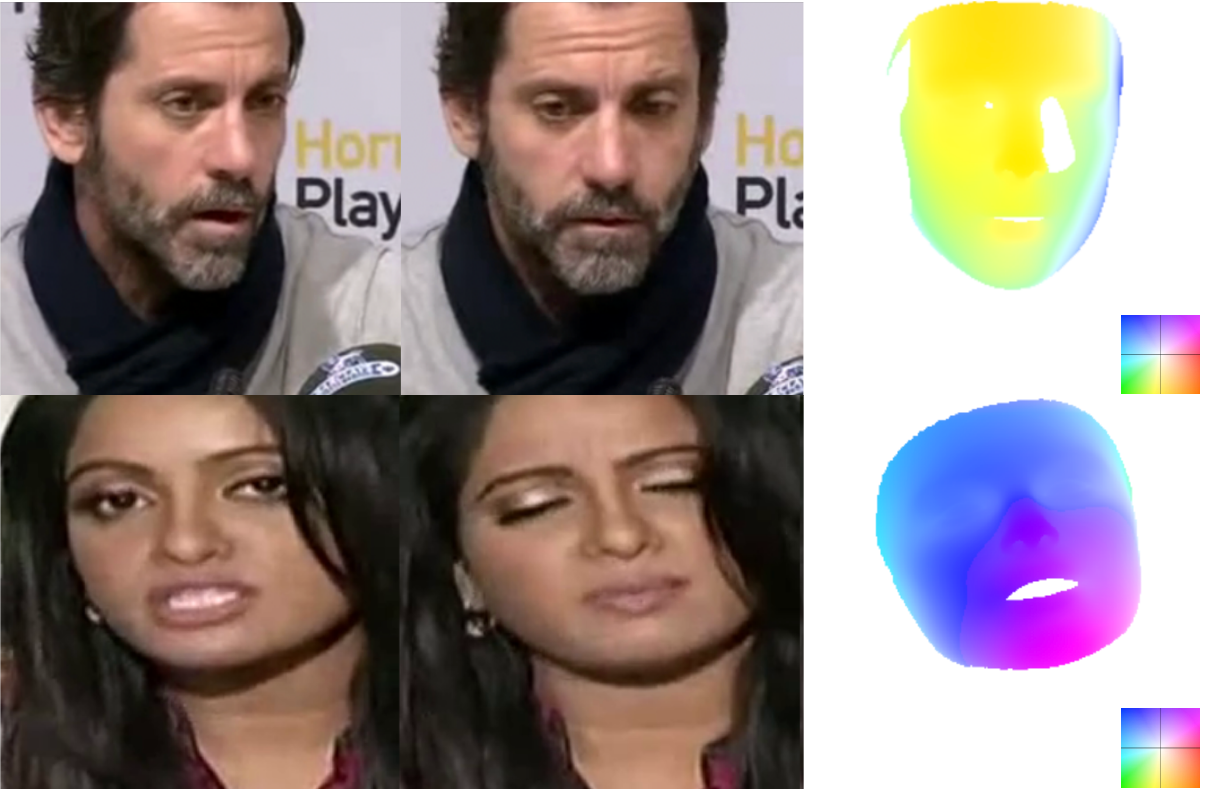}
		\end{minipage}
	}
	\vspace{-3mm}
	\caption{\small{(a) The illustration of barycentric coordinate interpolation of optical flow map. (b) Left: frames at time $t-1$; middle: frames at time $t$; right: generated flow maps $F^{t \rightarrow t-1}$ which warp frame $t$ back to frame $t-1$.}}
	\label{fig:flowmap3d}
\end{figure*}

%Typically, face swapping algorithms use image datasets such as FFHQ~\cite{karras2019style} for training considering the diversity of identities. However, image dataset struggles on the video face swapping task since it provides little temporal information. Besides, since the source and target images are sampled as two different people, it means that we have no ground truth in this case. For face reenactment, training pairs are taken from video datasets such as VoxCeleb2~\cite{chung2018voxceleb2}, usually sampled as consecutive frames from one individual video to form a self-supervised learning scenario.
%For video versions, three pairs of consecutive frames ($X_i^t, X_i^{t-1}$), ($X_p^t, X_p^{t-1}$) and ($X_e^t, X_e^{t-1}$) are sampled from one individual video clip. It means that $X_i^t$, $X_p^t$, and $X_e^t$ have the same identity, which forms self-supervision. For image versions,  $X_p^t$, $X_e^t$, $X_i^t$ are randomly sampled from image datasets with different identities. $X_p^t$ and $X_p^{t-1}$ are set to be the same frame, so are the expression images and identity images.
The whole dataset $V$ is expressed as $V=\left\{v_n\right\}_{n=1}^{N}$, where $N$ denotes the number of training videos. Each video contains a few frames, $v_{n}=\left\{f_{n, l}\right\}_{l=1}^{L_{n}}$, where $L_{n}$ denotes the number of frames in the $n$-th video $v_{n}$. Typically, we have two ways to sample training samples, namely intra-video sampling and inter-video sampling. The intra-video sampling selects training data $X_i^t$, $X_p^t$, and $X_e^t$ from the same video while inter-video sampling selects $X_i^t$, $X_p^t$, and $X_e^t$ from different videos. Intra-video sampling is a self-supervised learning scenario and provides ground-truth for training. Besides, optical flow warping based temporal loss can also be applied in this scenario. However, this sampling method is limited in self-driving swapping or reenactment and tends to fail in the more general cross-identity setting. For inter-video sampling, it provides more diverse samples for cross-identity training, making the framework more robust. However, it lacks ground-truth constraint and the temporal loss can not be applied either.  

In this paper, we demonstrate a joint training of face swapping and face reenactment with a novel dynamic training sample selection mechanism. %Specifically, we set the intra-video sampling ratio to be $\sigma \in [0,1]$. For each iteration, the intra-video sampling is selected if a random value $r$ is less than the threshold value $\sigma$, $r \textless \sigma$, and the sampling is conducted as:
For the intra-video sampling, we take three pairs of consecutive frames from one selected video. Specifically, the intra-video sampling ratio is set to be $\sigma \in [0,1]$ and conducted as:
\begin{align}
\begin{split}
X_p^t = f_{n, l_1},\ X_i^t &= f_{n, l_2},\ X_e^t = f_{n, l_3} \\
X_p^{t-1} = f_{n, l_1-1},\ X_i^{t-1} &= f_{n, l_2-1},\ X_e^{t-1} = f_{n, l_3-1}.
\end{split}
\end{align}
where $l_1$, $l_2$, $l_3$ are the index of random frames, $l_1, l_2, l_3 \in [2, L_n]$. %For the opposite case, $r \geq \sigma$, we take the inter-video sampling as follows.
For the opposite case, we take the inter-video sampling strategy where three pairs of consecutive frames are taken from three different videos.
\begin{align}
\begin{split}
X_p^t = f_{n_1, l_1},\ X_i^t &= f_{n_2, l_2},\ X_e^t = f_{n_3, l_3} \\
X_p^{t-1} = f_{n_1, l_1-1},\ X_i^{t-1} &= f_{n_2, l_2-1},\ X_e^{t-1} = f_{n_3, l_3-1}.
\end{split}
\end{align}
where $n_1$, $n_2$, $n_3$ are the index of random videos, $n_1, n_2, n_3 \in [1, N]$.

In this way, we use the intra-video sampling to provide ground truth supervision and the inter-video sampling with multiple identities to make the model more robust. Experiment results show that this multi-task training mechanism is efficient and benefits both tasks.

\begin{figure}[htb]
	\centering
	\includegraphics[width=0.9\linewidth]{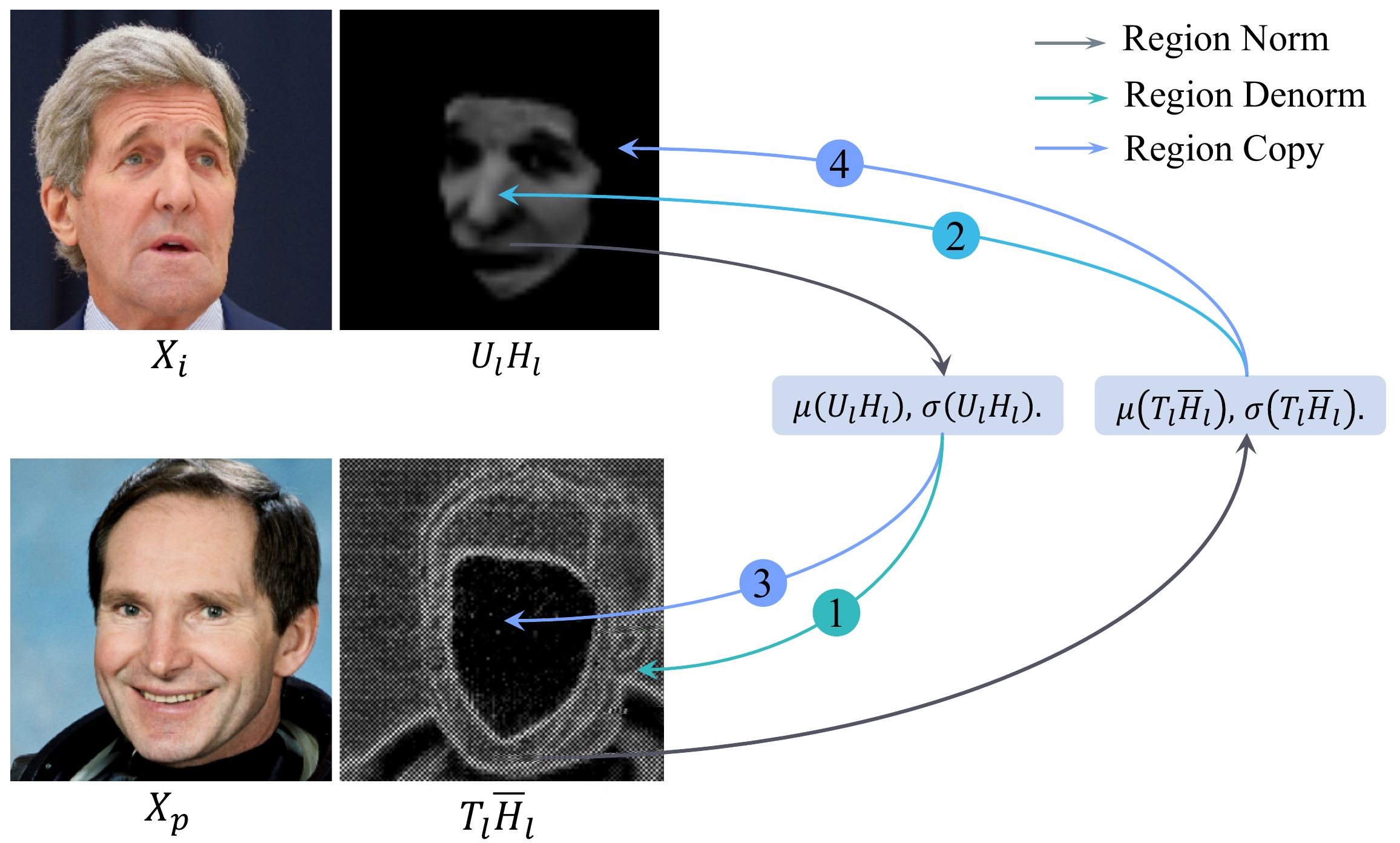}
    \vspace{-3mm}
	\caption{\small{The illustration of the region-aware conditional normalization layer. Both facial and non-facial areas are applied Batch Normalization first and then copied (same-regional retention terms) or denormalized (cross-regional transfer terms) to the corresponding areas. $T_l$ is an upsampled version of the appearance feature map $P$ and $H_l$ is a downsampled version of the facial mask $\hat{M}$. $U_l$ is the feature map in the upsampling process of $\mathbf{G}$. $\mu$ and $\sigma$ denotes the mean and standard deviation operations.}}
	\label{fig:BSN_diagram}
\end{figure}

\subsection{3D Disentangled Editing}\label{subsec:3D}
Since frames at time $t$ and $t-1$ share the same 3D reconstruction procedures, we ignore the superscript $t$ and $t-1$ here.
Given the identity image $X_i$, appearance image $X_p$ and expression image $X_e$, we estimate the 3DMM face shapes and the camera projection matrices using 3DDFA~\cite{zhu2017face}. For convenience, we set \(x = i, p, e\) to denote that we are dealing with \(X_i\), \(X_p\) and \(X_e\) respectively and the reconstructed 3D face can be expressed as follows:
%The 3DMM model explicitly decouples expression, identity, and pose information, which can later be recombined according to the requirement of different applications. 
%Mathematically, the reconstructed 3D face can be expressed as follows:
\begin{align}
\mathcal{S}^x = \overline {\mathcal{S}} + A_{\text{id}} \alpha_{\text{id}}^x + A_{\text{exp}} \alpha_{\text{exp}}^x ,
\end{align}
%where \(x = i, p, e\) denotes that we are dealing with \(X_i\), \(X_p\) and \(X_e\) respectively. To be specific,
where \(\mathcal{S}^x\) is a 3D face while \(\overline {\mathcal{S}}\) is the mean shape. \(A_{\text{id}}\) and \(A_{\text{exp}}\) are the principle axes derived from BFM~\cite{paysan20093d} and FaceWarehouse~\cite{cao2013facewarehouse}, respectively. $\alpha_\text{id}^x$ and $\alpha_\text{exp}^x$ are corresponding identity and expression coefficients. At the same time, the 3D reconstruction process also outputs the camera pose $C^x$ (i.e., the position, rotation and scale information). 
Next, we recombine the identity and expression coefficients to acquire the desired 3D face:
\begin{align}
\mathcal{S} = \overline {\mathcal{S}} + A_{\text{id}} \alpha_{\text{id}}^i + A_{\text{exp}} \alpha_{\text{exp}}^e,
\end{align}
where \(\mathcal{S}\) is the transformed 3D representation for the desired face. 
Meanwhile, we adopt $C^p$ as the selected pose.
Finally, we project the 3D face $\mathcal{S}$ with the texture map from $X_i$ onto the 2D image plane to obtain the rendering result:
\begin{align}
R = render(\mathcal{S}, T^i, C^p),
\end{align}
where \(render\) is a rasterization renderer with the Weak Perspective Projection according to the camera pose $C^p$, and $T^i$ denotes the texture map from $X_i$. The rendered face image $R$ contains the identity of \(X_i\), the expression of \(X_e\), and the pose of \(X_p\). For more details, please refer to \cite{zhu2017face}. Afterwards, the rendered face image $R$ is used to compute a binary facial region mask $\hat{M}$ with facial areas set to 1 and non-facial areas set to 0. Then we generate the appearance hint $M$. %Notably, we name it 'appearance hint' as it provides both the background and the pose information. 
Specifically, for the face swapping case, we set $M = X_p (1 - \hat{M})$ since swapping results require the same pose with $X_p$; for the face reenactment case, we set $M = X_i (1 - \hat{M})$ for similar reasons.

\subsection{Deep Blending Network}\label{subsec:GAN}
In this section, we describe the GAN-based Deep Blending Network which consists of an appearance embedder $\mathbf{E}$, an encoder-decoder structure generator $\mathbf{G}$, and a discriminator $\mathbf{D}$. Generally, for input frames at times $t$ and $t-1$, they share the same generation procedure, thus only the time $t$ version is explained and the superscript is also ignored in the following.

\noindent\textbf{Appearance Embedder \(\mathbf{E}(X_{p},M)\).}  It takes the concatenation of the appearance frame $X_{p}$ and the appearance hint image $M$, and maps these inputs into a low-resolution appearance feature map denoted as $P$. 

\noindent\textbf{Generator \(\mathbf{G}(R, P, \hat{M})\).} The rendered 3D face $R$ and the appearance embedding feature map $P$ are input to $\mathbf{G}$ to synthesize the video frame $Y$. %We build $\mathbf{G}$ using the widely adopted image-to-image translation network~\cite{johnson2016perceptual} but replace the instance normalization with a novel bidirectional region-aware normalization layer (BRN), which is an enhanced version of AdaIN~\cite{huang2017arbitrary}.% add in supplementary 
For the better fusion result, we propose a novel region-aware conditional normalization layer (RCN) layer tailored for our network as presented in Fig~\ref{fig:BSN_diagram}. In general, RCN computes the mean and variance for facial and non-facial areas respectively and conducts the denormalization in both the same-regional retention or cross-regional transfer way. Specifically, for the same-region retention term, we firstly apply batch normalization and then copy the region features to the corresponding parts. For the cross-regional term, we transfer the region statistic to the counterpart with a relatively small weight.

For each feature map $U_l$ in the upsampling process of $\mathbf{G}$, $ 1 \leq l \leq L$, where $L$ is the number of upsampling layers, RCN is expressed as follows. %For the sake of convenience, we ignore the superscript $t$ here:
\begin{equation}
\begin{split}
\text{RCN}(U_l, T_l, H_l) & = \underbrace{\alpha_l \cdot \mathcal{T}(U_l H_l, T_l \overline{H}_l)  + \beta_l \cdot \mathcal{T}(T_l\overline {H_l}, U_l H_l)}_\text{cross-regional transfer term}  \\
& +  \underbrace{ (1 - \alpha_l) \cdot U_l H_l  + (1 - \beta_l) \cdot T_l \overline {H}_l}_\text{same-regional retention term}  \label{eqn:RCN}
\end{split}
\end{equation}
and $\mathcal{T}(\cdot, \cdot)$ follows AdaIN  as: %\scriptsize $\mathcal{T}(A, B) = \sigma(B) \Big(\frac{A - \mu(A)}{\sigma (A)}\Big) + \mu(B), $
\begin{align}
\mathcal{T}(A, B) = \sigma(B) \Big(\frac{A - \mu(A)}{\sigma (A)}\Big) + \mu(B), 
\end{align}
where $T_l$ is an upsampled version of the appearance feature map $P$ and $H_l$ is a downsampled version of the facial mask $\hat{M}$. $\overline{H}_l = 1 - H_l$ indicates the non-facial area. Two learnable parameter vectors $\alpha_l$ and $\beta_l$ are adopted for balance. \(\alpha_l \in \mathbb{R}^{1\times 1\times c}\) and \(\beta_l \in \mathbb{R}^{1\times 1\times c} \) share the same channel as input $U_l$. Each element \(\alpha_i \in [0,1]\) and \(\beta_i \in [0,1], i \in [1, c]\), and the initialization values are set to 0.8 and 0.1, respectively. %\textcolor{red}{The output is composed of the same-regional retention term and the cross-regional transfer term. Specifically, for the same-region retention term, we firstly apply batch normalization and then copy the region features to the corresponding parts. For the cross-regional term, we transfer the region statistic to the other region.}

\textbf{\textit{Why should the denormalization process be limited in the specific region?}} AdaIN~\cite{huang2017arbitrary} treats elements in the same channel equally and aligns the channel-wise mean and variance. SPADE~\cite{park2019semantic} predicts pixel-level affine parameters and applies spatial-aware denormalization. However, in our case, since Generator $\mathbf{G}$ and Appearance Embedder $\mathbf{E}$ encode the facial and non-facial areas respectively, restricting the computation in specific areas gives more accurate statistics and avoids the interference of meaningless feature regions. \textbf{\textit{Why should we need a cross-regional transfer term?}} This cross-regional transfer term applies the facial statistic to the non-facial areas with a relatively small weight and vice versa. This mechanism makes the two regions adjust to each other and when training end-to-end, it leads to more context-harmonious results.
%As shown in Fig~\ref{fig:BSN_diagram}, in addition to using $B_p$ to regulate $F_i$, BRN also uses $F_i$ to regulate $B_p$. In this way, the feature statistic of the facial can be applied to the background too and generate more context-harmonious results when training end-to-end. \textbf{\textit{Why should the non-facial statistic be applied to both the facial and non-facial areas?}} In Fig~\ref{fig:BSN_diagram}, the mean and variance value of the background area of $P^t$ are used to denormalized both the facial and non-facial areas, which leads to the more consistent fusion edge.
%In our case, however, the embedding feature maps $P^t$ is spatial-aware and appearance-related, which should mainly be applied outside facial areas. Therefore, we take one step further to develop BSN which utilizes the facial region mask $\hat{M}^t$ as a prior prompt. The illustration of BSN is presented in Fig~\ref{fig:BSN_diagram}, showing the bidirectional denormalization. Specifically, 

\noindent\textbf{Discriminator \(\mathbf{D}(Y, X_i)\).} Following pix2pixHD~\cite{wang2018high}, we adopt the same multi-scale discriminator and loss functions, except that we replace the least squared loss term~\cite{mao2017least} by the hinge loss term~\cite{lim2017geometric}.

\setlength{\tabcolsep}{4pt}
\begin{table}[t]
	\begin{center}
		\caption{\small{Comparison results for face swapping and face reenactment on VoxCeleb2 with $\sigma$ set to 0.5. Note that SSIM for face reenactment is calculated in a self-driving scenario.}}
		\vspace{-2mm}
		\label{table:swapReenactCompare}
		%\begin{tabular}{llllllll}
		\resizebox{\linewidth}{!}{
			\begin{tabular}{ccccccccc}
				%\hline\noalign{\smallskip}
				%\toprule[1pt]
				\toprule
				Mode & Model   & FID \(\downarrow\) & SSIM \(\uparrow\) & $E_{\text{id}}$\(\downarrow\) & $E_{\text{pose}}$\(\downarrow\)  & $E_{\text{exp}}$ \(\downarrow\) \\
				\noalign{\smallskip}
				%\hline
				%\toprule[1pt]
				\midrule
				\noalign{\smallskip}
				\multirow{3}*{\scriptsize
					Swapping} & FaceShifter  & 52.1 & - & 1.31 & \scriptsize(4.31, 3.26, 1.13) & 4.43 \\ % (4.31, 3.26, 1.13) 
				~ & FSGAN & 45.9 & - & 0.89  & \scriptsize(2.27, 3.21, 0.76) & 3.23 \\ % (2.27, 3.21, 0.76) 
				~ &  UniFaceGAN & \textbf{34.1} & -   & \textbf{0.26} & \scriptsize(\textbf{1.12}, \textbf{1.31}, \textbf{0.65}) & \textbf{2.31} \\ %(\textbf{1.12}, \textbf{1.31}, \textbf{0.65})
				\midrule
				\multirow{6}*{\scriptsize
					Reenactment} & X2face  & 52.1 & 0.05  & 0.29 & \scriptsize(2.31, 1.26, 0.90) & 4.43 \\ %(2.31, 1.26, 0.90)
				~ & FSGAN  & 32.4 & 0.23   & \textbf{0.10}  & \scriptsize(1.43, 0.75, 0.90) & 3.58  \\ % (1.43, 0.75, 0.90)
				~ & FTH  & 43.1 & 0.18 & 0.43  & \scriptsize(1.50, 1.23, 0.81) & 5.67  \\ %(1.50, 1.23, 0.81)
				~ & FOMM  & 28.2 & 0.59   & 0.12  & \scriptsize(0.76, 0.54, 0.12) & 4.02  \\  %(0.76, 0.54, 0.12) 
				~ & Fast Bi-layer  & 45.8 & 0.22   & 0.14  & \scriptsize(0.23, 0.34, 0.12) & 3.87  \\ % (0.23, 0.34, 0.12)
				~ &  UniFaceGAN  & \textbf{24.4}  & \textbf{0.82}  & 0.12 & \scriptsize(\textbf{0.21}, \textbf{0.25}, \textbf{0.12}) & \textbf{3.12}  \\ %(\textbf{0.21}, \textbf{0.25}, \textbf{0.12}) 
				%\hline
				%\toprule[1pt]
				\bottomrule
			\end{tabular}
		}
	\end{center}
\end{table}

\setlength{\tabcolsep}{4pt}
\begin{table}[t]
	\begin{center}
		\caption{\small{Ablation studies for Dynamic Training Sample Selection mechanism.}}
		\vspace{-3mm}
		\label{table:DTSSAblation}
		%\begin{tabular}{llllllll}
		\resizebox{\linewidth}{!}{
			\begin{tabular}{cccccccc}
				%\hline
				\noalign{\smallskip}
				%\toprule[1pt]
				\toprule
				Mode   & $\sigma$ & FID\(\downarrow\) & SSIM\(\uparrow\) & $E_{\text{id}}$\(\downarrow\) & $E_{\text{pose}}$\(\downarrow\)  & $E_{\text{exp}}$\(\downarrow\) & $E_{\text{tmp}}$\(\downarrow\) \\
				\noalign{\smallskip}
				%\hline
				%\toprule[1pt]
				\midrule
				\noalign{\smallskip}
				\multirow{6}*{Swapping} & \cellcolor{mygray}0.0  & \cellcolor{mygray}34.5 & \cellcolor{mygray}- & \cellcolor{mygray}0.28 & \scriptsize\cellcolor{mygray}(1.31, 1.33, \textbf{0.42}) & \cellcolor{mygray}\textbf{2.18} & \cellcolor{mygray}4.23 \\
				~ &  0.2 & 34.3 & -   & 0.26 & \scriptsize(1.12, \textbf{1.30}, 0.65) & 2.20 & 4.01\\
				~ &  0.4 & 34.1 & -   & 0.27 & \scriptsize(\textbf{1.11}, 1.33, 0.65) & 2.23 & 3.74\\
				~ &  0.5 & \textbf{34.1} & -   & \textbf{0.26} & \scriptsize(1.12, 1.31, 0.65) & 2.31 & 3.54\\
				~ &  0.6 & 36.3 & -   & 0.51 & \scriptsize(1.18, 1.31, 0.64) & 2.48 & 3.49\\
				~ &  0.8 & 38.2 & -   & 1.02 & \scriptsize(1.20, 1.32, 0.65) & 2.56 & \textbf{3.45}\\
				%\hline
				\midrule
				\multirow{6}*{Reenactment} & 0.2  & 25.1 & 0.80 & 0.31 & \scriptsize(0.42, 0.38, 0.12) & 4.01 & 4.03 \\
				~ &  0.4 & 24.8 & 0.81   & 0.26 & \scriptsize(0.40, 0.34, \textbf{0.11}) & 3.30 & 3.82\\
				~ &  0.5 & \textbf{24.4} & 0.82 & \textbf{0.12} & \scriptsize(\textbf{0.21}, \textbf{0.25}, 0.12) & \textbf{3.12} & 3.65\\
				~ &  0.6 & 26.1 & 0.85   & 0.29 & \scriptsize(0.27, 0.31, 0.12) & 3.92 & 3.62\\
				~ &  0.8 & 26.5 & 0.86   & 0.32 & \scriptsize(0.30, 0.32, 0.15) & 4.08 & 3.56\\
				~ &  \cellcolor{mygray}1.0 & \cellcolor{mygray}26.9 & \cellcolor{mygray}\textbf{0.87}  & \cellcolor{mygray}0.33 &\scriptsize\cellcolor{mygray}(0.31, 0.33, 0.15) & \cellcolor{mygray}4.11 & \cellcolor{mygray}\textbf{3.25}\\
				%\toprule[1pt]
				\bottomrule
			\end{tabular}
		}
	\end{center}
\end{table}

\begin{figure}[t] % select 2
%\begin{figure*}[t] % select 2
	\centering
	\subfigure[]{
		\begin{minipage}[b]{0.16\linewidth}
			\includegraphics[width=1.25\linewidth,height=1.25\linewidth]{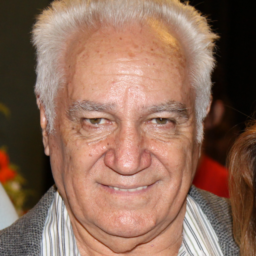}
			\includegraphics[width=1.25\linewidth,height=1.25\linewidth]{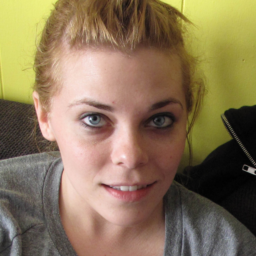}
		\end{minipage}
	}
	\subfigure[]{
		\begin{minipage}[b]{0.16\linewidth}
			\includegraphics[width=1.25\linewidth,height=1.25\linewidth]{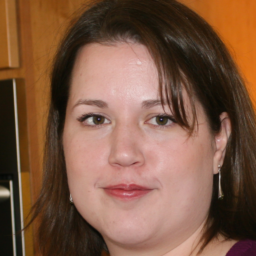}
			\includegraphics[width=1.25\linewidth,height=1.25\linewidth]{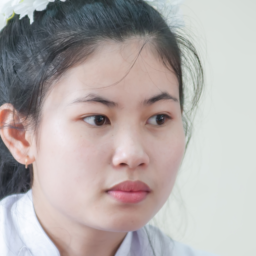}
		\end{minipage}
	}
	\subfigure[]{
	\begin{minipage}[b]{0.16\linewidth}
		\includegraphics[width=1.25\linewidth,height=1.25\linewidth]{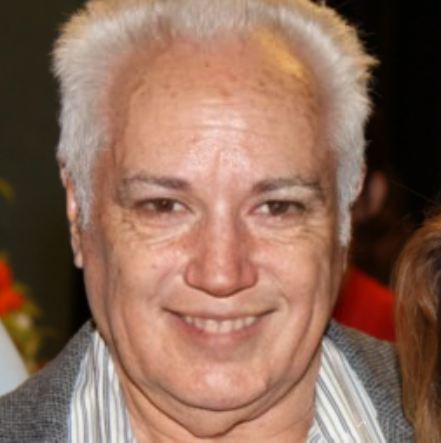}
		\includegraphics[width=1.25\linewidth,height=1.25\linewidth]{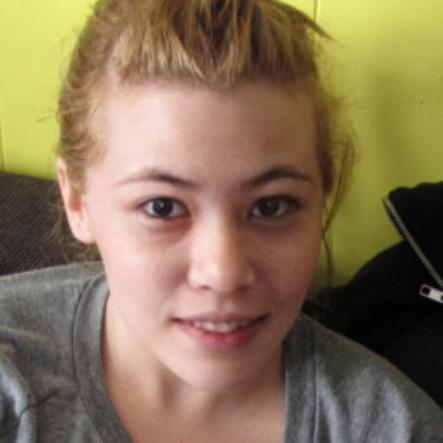}
	\end{minipage}
	} 
	\subfigure[]{
		\begin{minipage}[b]{0.16\linewidth}
			\includegraphics[width=1.25\linewidth,height=1.25\linewidth]{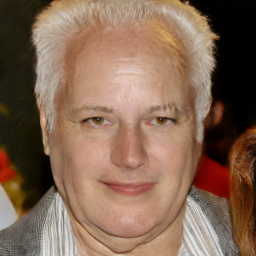}
			\includegraphics[width=1.25\linewidth,height=1.25\linewidth]{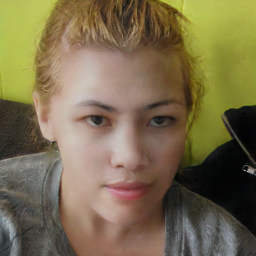}
		\end{minipage}
	} 
	\subfigure[]{
		\begin{minipage}[b]{0.16\linewidth}
			\includegraphics[width=1.25\linewidth,height=1.25\linewidth]{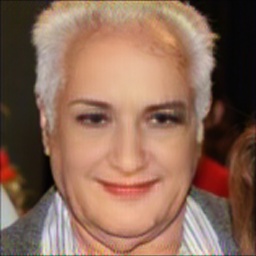}
			\includegraphics[width=1.25\linewidth,height=1.25\linewidth]{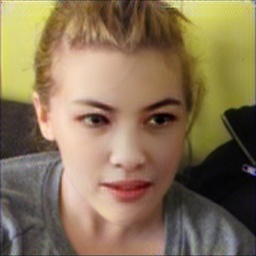}
		\end{minipage}
	}
	%\subfigure[]{
	%	\begin{minipage}[b]{0.12\linewidth}
	%		%\includegraphics[width=1.2\linewidth,height=1.2\linewidth]{results_swap/654_swap.png}
	%		\includegraphics[width=1.2\linewidth,height=1.2\linewidth]{results_swap/110_swap.png}
	%		%\includegraphics[width=1.2\linewidth,height=1.2\linewidth]{results_swap/1210_swap.png}	
	%		\includegraphics[width=1.2\linewidth,height=1.2\linewidth]{results_swap/1861_swap.png}	
	%		%\includegraphics[width=1.2\linewidth,height=1.2\linewidth]{results_swap/1292_swap.png}
	%		%\includegraphics[width=1.2\linewidth,height=1.2\linewidth]{results_swap/1892_swap.png}
	%	\end{minipage}
	%}
	\vspace{-3mm}
	\caption{\small{Comparison results of face swapping. (a) Input appearance images $X_p$. (b) Input identity images $X_i$. Results of: (c) UniFaceGAN(Ours); (d) FaceShifter; (e) FSGAN.}}
	\label{fig:swapCompare}
%\end{figure*}
\end{figure}
\subsection{Loss Function}\label{subsec:loss}
\noindent\textbf{Appearance Preserving Loss.} The generated facial image $Y^t$ should have the same appearance information as $X_p^t$ and the appearance preserving loss is measured in $L_1$ norm as follows:
\begin{align}
\mathcal{L}_{{app}} = || \mathbf{E}(Y^{t})- \mathbf{E}(X_p^t) \|_{1},
\end{align}
where $\mathbf{E}$ is the appearance embedder described in Section~\ref{subsec:GAN}.

\noindent\textbf{Reconstruction Loss.}  When the dynamic training sample selection mechanism chooses the intra-video sampling,  we have ground truths for supervised training, since $X_i^t$ and $X_p^t$ are from the same person. Then we have the reconstruction loss as follows:
\begin{equation}
\mathcal{L}_{{rec}} = || Y^{t}- X_i^t \|_{1}.
\end{equation}
Note that for the inter-video sampling case, $\mathcal{L}_{{rec}}$ is set to 0.

\noindent\textbf{Adversarial Loss.} For $\mathcal{L}_{{adv}}$, we use a multi-scale adversarial loss~\cite{wang2018high} on the downsampled output image to enforce photo-realistic results:
\begin{align}
\mathcal{L}_{{adv}} = \frac{1}{K} \sum_{k=1}^{K} \mathbb{E}_{{X_{i,k}^t}}[\log \mathbf{D}(X_{i,k}^t)]+\mathbb{E}_{{Y_k^t}}[\log (1-\mathbf{D}(Y_k^t))],
\end{align}
where $X_{i,k}^t$ and $Y_k^t$ are the downsampled versions of $X_i^t$ and $Y^t$, respectively, and $K$ is the total number of scale versions.

\noindent\textbf{3D Temporal Loss.} Insipred by \cite{dong2018supervision}, we introduce a flow-based 3D temporal loss to alleviate the inter-frame flicker artifacts. The main difficulty lies in how to generate the dense optical flow map between two consecutive frames. Here, we propose a 3D-based optical flow extraction method, which avoids the use of flow estimation networks~\cite{ilg2017flownet}.

Thanks to 3DMM, the optical flow values for reconstructed vertices can be easily obtained by directly subtracting 3D coordinates between adjacent frames. To obtain the dense flow map, we propose to conduct barycentric coordinate interpolation for each pixel in the 2D pixel domain.  As shown in Fig.~\ref{fig:flowmap3d}(a), for a query point $q$ with location $(i,j)$, we find the triangle $T \in \tau$ with its $x$-$y$ plane projection version containing $q$. Let $V_{1}^{t}$, $V_{2}^{t}$, $V_{3}^{t} \in \mathbb{R}^{3}$ be the vertices of $T$. We compute the barycentric coordinates  $\lambda_{1}^{t},\lambda_{2}^{t},\lambda_{3}^{t}$ of $q$, with respect to $T$. Therefore, the flow value at position $(i, j)$ is calculated as follows:
\begin{align}
W_{i, j}^{t \rightarrow t-1}=\sum_{k=1}^{3} \lambda_{k}^{t}\left(V_{k}^{t}-V_{k}^{t-1}\right).
\end{align}
Then we need to obtain visibility maps $S^{t-1}, S^t$. Let $Z^{t-1}$ and $Z^{t}$ be the depth buffer generated by 3DDFA. For frame $t$, we interpolate the depth of $q$ to get the coordinate of $Q$:
\begin{align}
Q^{t} =  (i, j, Q^{t}_z)  = (i, j, \sum_{k=1}^{3} \lambda_{k}^{t} z_{k}^{t}),
\end{align}
where $Q^{t}_z$ denotes the $z$ component of $Q^{t}$. Since the depth buffer denotes the minimum visible \textit{z} value of each coordinate of the 2D pane to project on, the visibility map $S^t$ value at $(i,j)$ is given as:
\begin{align}
S^t_{i,j}=\left\{\begin{array}{ll}
 1 & \text { if } {Q^t_z} >= Z^{t}_{i,j}\\
0 & \text { otherwise }
\end{array}\right.
\end{align}
where $Z^{t}_{i,j}$ means the depth buffer value of $Z^{t}$ at point $(i,j)$.
For frame $t-1$, we compute the coordinate of $Q^{t-1}$ as follows:
\begin{align}
Q^{t-1} =  ({Q^{t-1}_{x}}, {Q^{t-1}_{y}}, {Q^{t-1}_{z}})  = Q^{t} - W_{i, j}^{t \rightarrow t-1},
%Q^{t-1}_{i,j}  = Q^{t}_{i,j} - W_{i, j}^{t \rightarrow t-1}
\end{align}
Since $Q^{t}$ and $W_{i, j}^{t \rightarrow t-1}$ are known, we can obtain $Q^{t-1}_{x}$, $Q^{t-1}_{y}$,  $Q^{t-1}_{z}$ (\textit{i.e.} $x$, $y$, $z$ components of $Q^{t-1}$). Then the calculation of visibility map $S^{t-1}$ for frame $t-1$ at position $({Q^{t-1}_{x}, Q^{t-1}_{y}})$  is similar to $S^t$.
\begin{align}
S^{t-1}_{Q^{t-1}_{x}, Q^{t-1}_{y}}=\left\{\begin{array}{ll}
1 & \text { if } Q^{t-1}_{z} >= Z^{t-1}_{Q^{t-1}_{x}, Q^{t-1}_{y}}\\
0 & \text { otherwise }
\end{array}\right.
\end{align}
Finally, the optical flow map between two consecutive frames is given as follows:
\begin{align}
F_{i, j}^{t \rightarrow t-1} = W_{i, j}^{t \rightarrow t-1} \cdot S^t_{i,j} \cdot S^{t-1}_{Q^{t-1}_{x}, Q^{t-1}_{y}}.
\label{con:flowCal}
\end{align}
Therefore, we define the 3D temporal loss between $Y^t$ and $Y^{t-1}$ in the format of mean squared error:
\begin{align}
\mathcal{L}_{tmp} = || Y^{t-1}-warp\left(Y^{t}, F^{t\rightarrow t-1}\right) \|_{2},\label{con:calculateTMP}
\end{align}
where $warp(Y^{t}, F^{t \rightarrow t-1})$ is the warping function that warps the output at time $t$ back to time $t-1$. The overall loss function is thus reached as :
\begin{align}
%L = \lambda_i L_{\text{id}} + \lambda_r L_{\text{rec}}  + \lambda_b L_{\text{bg}} + \lambda_a L_{\text{adv}} + \lambda_t L_{\text{tmp}},
 \mathcal{L} =  \arg \min _{\mathbf{G},\mathbf{E}} \max _{\mathbf{D}} \alpha \mathcal{L}_{{adv}}  + \beta \mathcal{L}_{{app}}  + \gamma \mathcal{L}_{{rec}} + \lambda \mathcal{L}_{tmp},
\end{align}
where  $\alpha$, $\beta$,  $\gamma$ and $\lambda$ are the weights to balance different terms, which are set to 10, 1, 10 and 5 in our experiments, respectively.
\section{Experiments}
\label{others}

\begin{figure*}[t]
	\centering
	\subfigure[]{
		\begin{minipage}[b]{0.1\linewidth}
			\includegraphics[width=1.2\linewidth,height=1.2\linewidth]{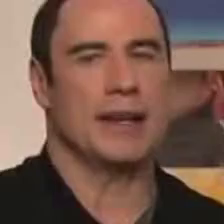}
			\includegraphics[width=1.2\linewidth,height=1.2\linewidth]{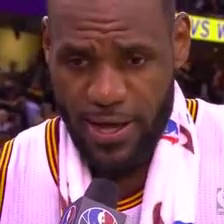}
		\end{minipage}
	}
	\subfigure[]{
		\begin{minipage}[b]{0.1\linewidth}
			\includegraphics[width=1.2\linewidth,height=1.2\linewidth]{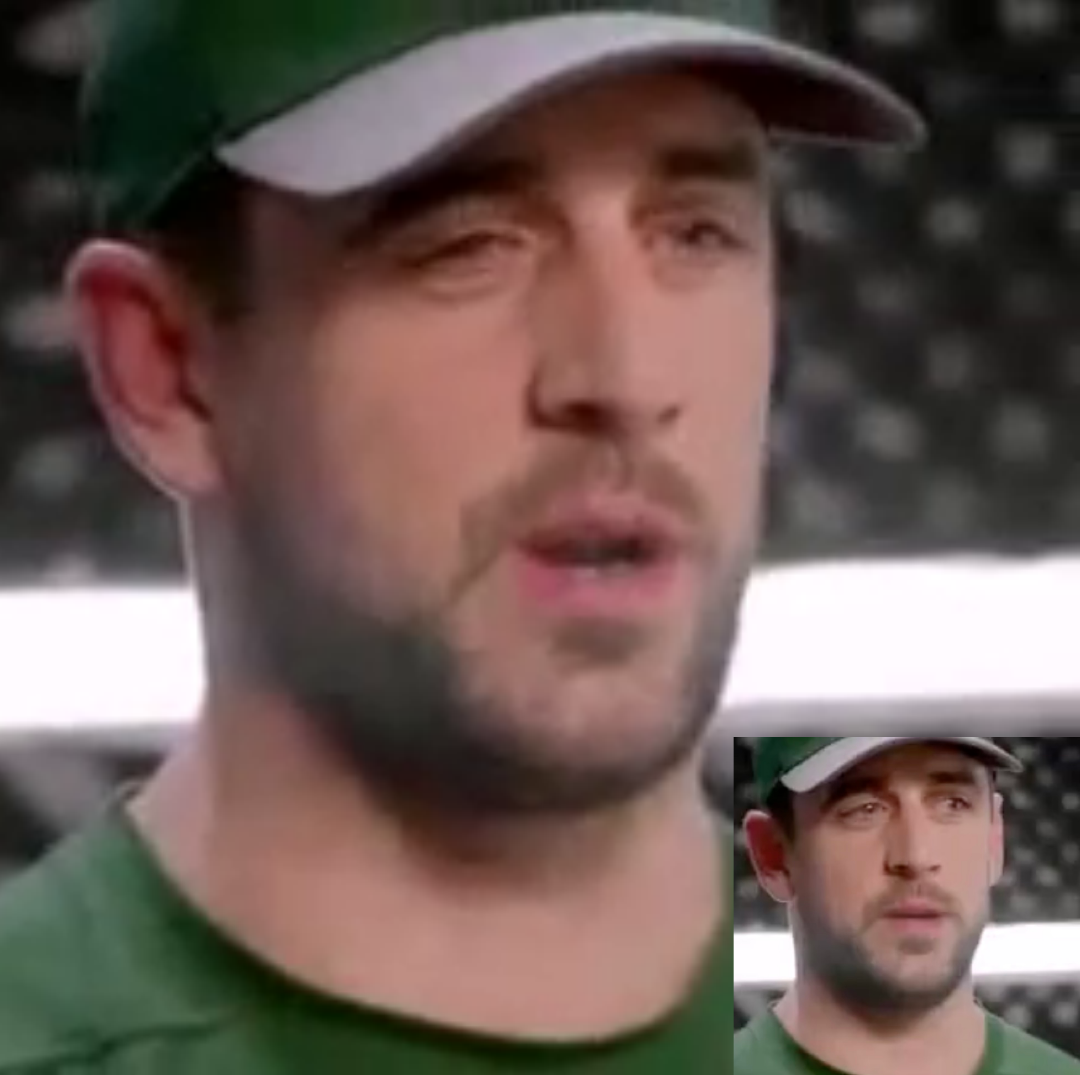}
			\includegraphics[width=1.2\linewidth,height=1.2\linewidth]{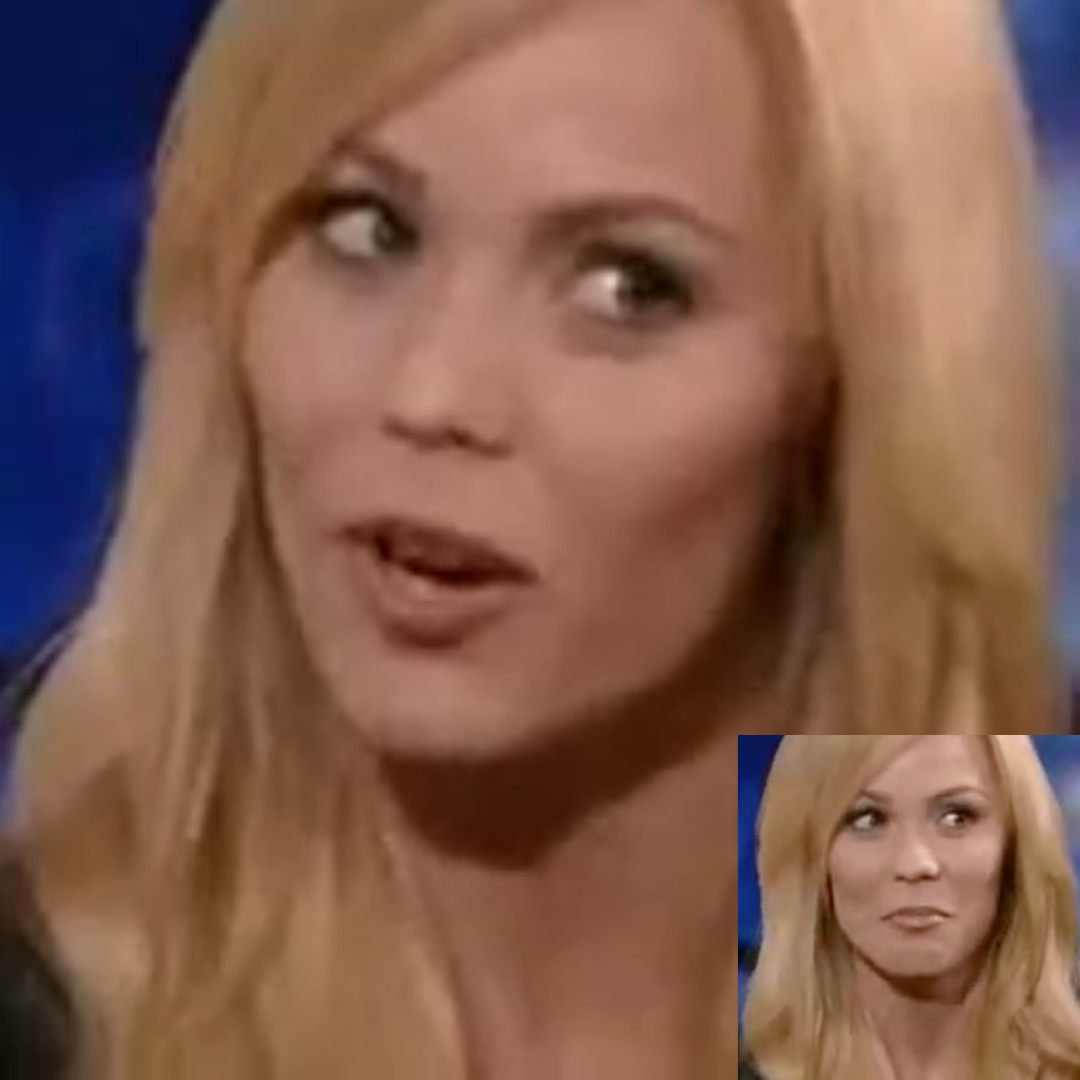}
		\end{minipage}
	}
	\subfigure[]{
		\begin{minipage}[b]{0.1\linewidth}
			\includegraphics[width=1.2\linewidth,height=1.2\linewidth]{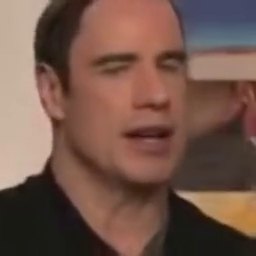}
			\includegraphics[width=1.2\linewidth,height=1.2\linewidth]{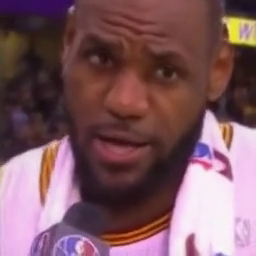}
		\end{minipage}
	} 
	\subfigure[]{
		\begin{minipage}[b]{0.1\linewidth}
			\includegraphics[width=1.2\linewidth,height=1.2\linewidth]{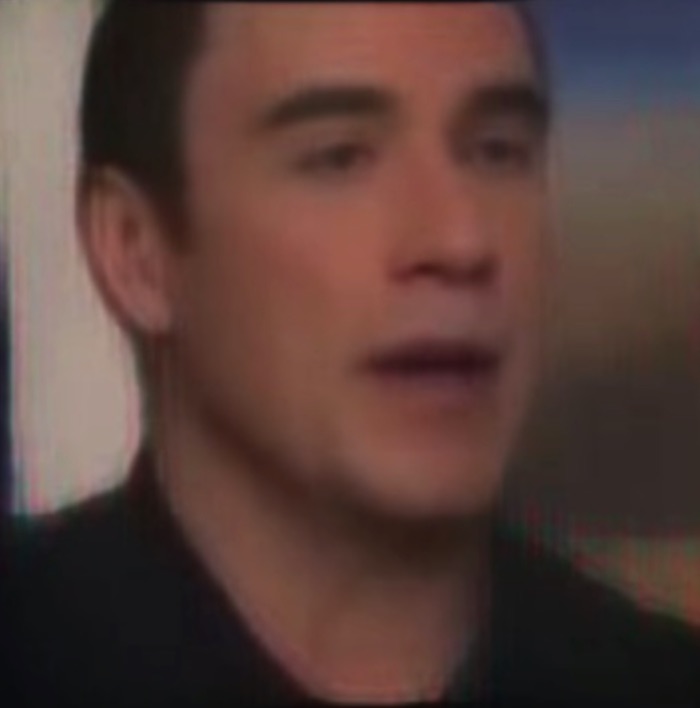}
			\includegraphics[width=1.2\linewidth,height=1.2\linewidth]{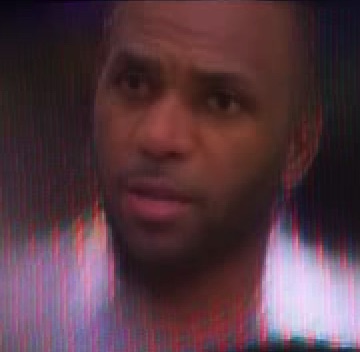}
		\end{minipage}
	}
	\subfigure[]{
		\begin{minipage}[b]{0.1\linewidth}
			\includegraphics[width=1.2\linewidth,height=1.2\linewidth]{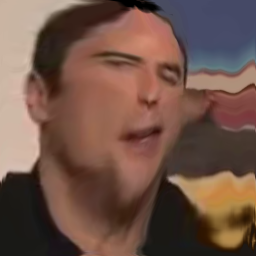}
			\includegraphics[width=1.2\linewidth,height=1.2\linewidth]{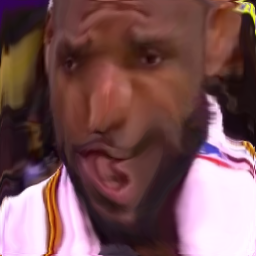}
		\end{minipage}
	}
	\subfigure[]{
		\begin{minipage}[b]{0.1\linewidth}
			\includegraphics[width=1.2\linewidth,height=1.2\linewidth]{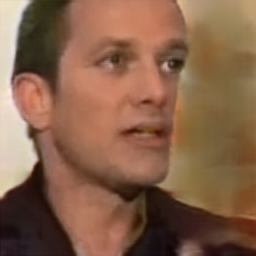}
			\includegraphics[width=1.2\linewidth,height=1.2\linewidth]{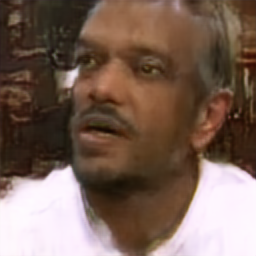}
		\end{minipage}
	}
	\subfigure[]{
		\begin{minipage}[b]{0.1\linewidth}
			\includegraphics[width=1.2\linewidth,height=1.2\linewidth]{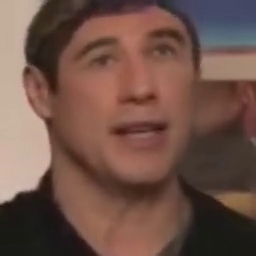}
			\includegraphics[width=1.2\linewidth,height=1.2\linewidth]{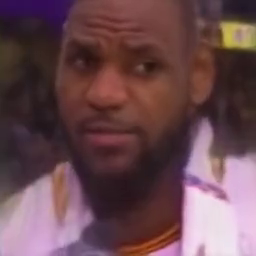}
		\end{minipage}
	}
	\subfigure[]{
		\begin{minipage}[b]{0.1\linewidth}
			\includegraphics[width=1.2\linewidth,height=1.2\linewidth]{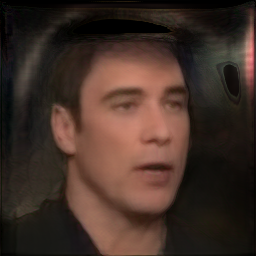}
			\includegraphics[width=1.2\linewidth,height=1.2\linewidth]{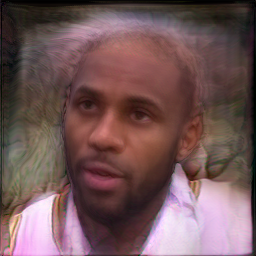}
		\end{minipage}
	}
	\vspace{-3mm}
	\caption{\small{Comparison results of face reenactment. (a) Input identity images. (b) Input appearance images. On the right corner is the first frame of the appearance video. It shares the similar pose with the identity image and we choose it as $\mathbf{D}_{1}$ for FirstOrder inference. (c) UniFaceGAN (Ours). (d) FSGAN. (e) X2face. (f) FTH (Few-shot T. Heads). (g) FOMM (First Order Motion Model). (h) Fast Bi-layer.}}
	\label{fig:reenactmentCompare}
\end{figure*}
\vspace{-2mm}

\begin{figure}[htb]
	\centering
	\includegraphics[width=0.9\linewidth]{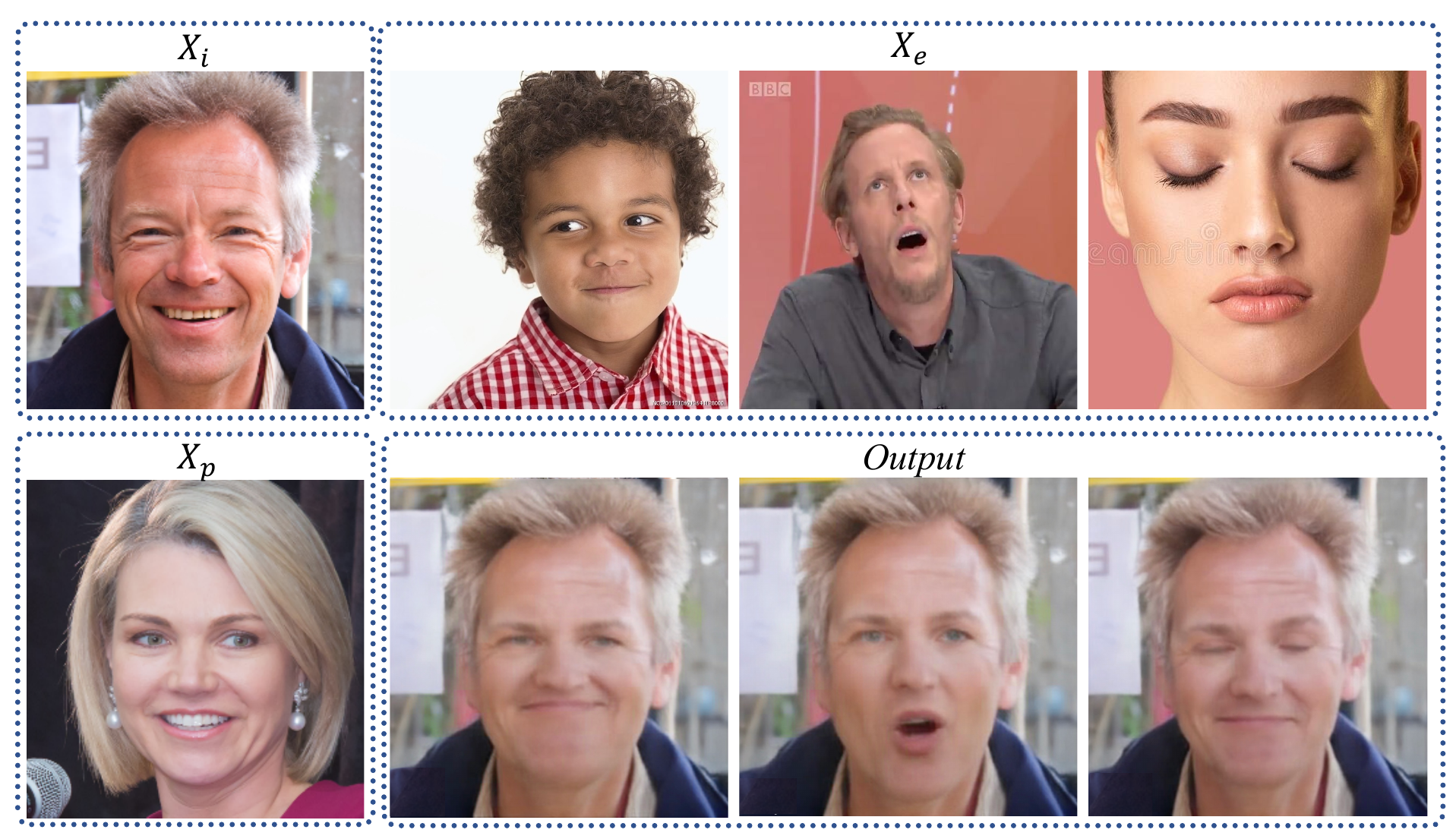}
	\vspace{-2mm}
	\caption{\small{The fully disentangled manipulation task.}}
	\label{fig:faceManipulation_new}
\end{figure}

\subsection{Experiment Details}
%In this paper, the FFHQ~\cite{karras2019style} and Voxceleb2~\cite{chung2018voxceleb2} datasets are used as image and video datasets, respectively. FFHQ consists of 70k high-quality images, containing plentiful variations in terms of age, ethnicity and background, while Voxceleb2 covers over a million utterances from over 6,000 speakers. The image dataset ratio $\sigma$ is set to 0.5.
In this paper, the Voxceleb2~\cite{chung2018voxceleb2} dataset is adopted to conduct the multi-task training. Voxceleb2 is among the largest available video datasets, which covers over a million utterances from over 6,000 speakers. The intra-video sampling rate $\sigma$ is set to 0.5. Quantitative comparisons are conducted with respect to five metrics: Frechet-inception distance (FID)~\cite{heusel2017gans}, structured similarity (SSIM)~\cite{wang2004image}, identity error $E_{\text{id}}$, pose error $E_{\text{pose}}$, and expression error $E_{\text{exp}}$.
%The networks are optimized using Adam optimizer with the learning rate set to \(1 \times 10^{-4}\) for the generator and \(4 \times 10^{-4}\) for the discriminator, respectively. Our framework is implemented using PyTorch and the training is carried out on 8 NVIDIA P40 GPUs with batch size 24.
More about the training details, evaluation metrics and the network architecture are available in the supplementary material.
%FID is computed between $Y^t$ and $X_i^t$ while SSIM is only computed for the self-driving face reenactment task. We input $X_i^t$ and $Y^t$ to a face recognition model~\cite{wang2018cosface} to extract identity vectors, and $E_{\text{id}}$ is evaluated in the cosine similarity. To measure pose accuracy, we calculate $E_{\text{pose}}$ using the Euclidean distance between the Euler angles of $X_p^t$ and $Y^t$ extracted by~\cite{ruiz2018fine}. Similarly, $E_{\text{exp}}$ is Euclidean distance between the 2D landmarks of $X_e^t$ and $Y^t$ detected by~\cite{zhang2016joint}. Specifically, we compute all metrics based on 100 videos from VoxCeleb2.

%The architecture of Appearance Embedder is based on pix2pix~\cite{isola2017image} with the input channel set to 6. In the encoding process, each convolution layer is followed by a leaky ReLU with factor 0.1 and an instance normalization layer. We construct our generator \(G\) using the image-to-image translation architecture proposed by Johnson et. al.~\cite{johnson2016perceptual} with the downsampling and upsampling layers replaced by residual blocks. The proposed region-aware conditional normalization layer (RCN) is only applied in the upsampling layer while the regular (non-adaptive) instance normalization layers are used in the downsampling blocks; besides, self-attention blocks are also used at \(64 \times 64\) resolution in the upsampling part of the generator. We apply seven downsampling residual blocks and the corresponding upsampling blocks. 

\subsection{Comparison Results}
\begin{figure}[t]
	\centering
	\includegraphics[width=\linewidth]{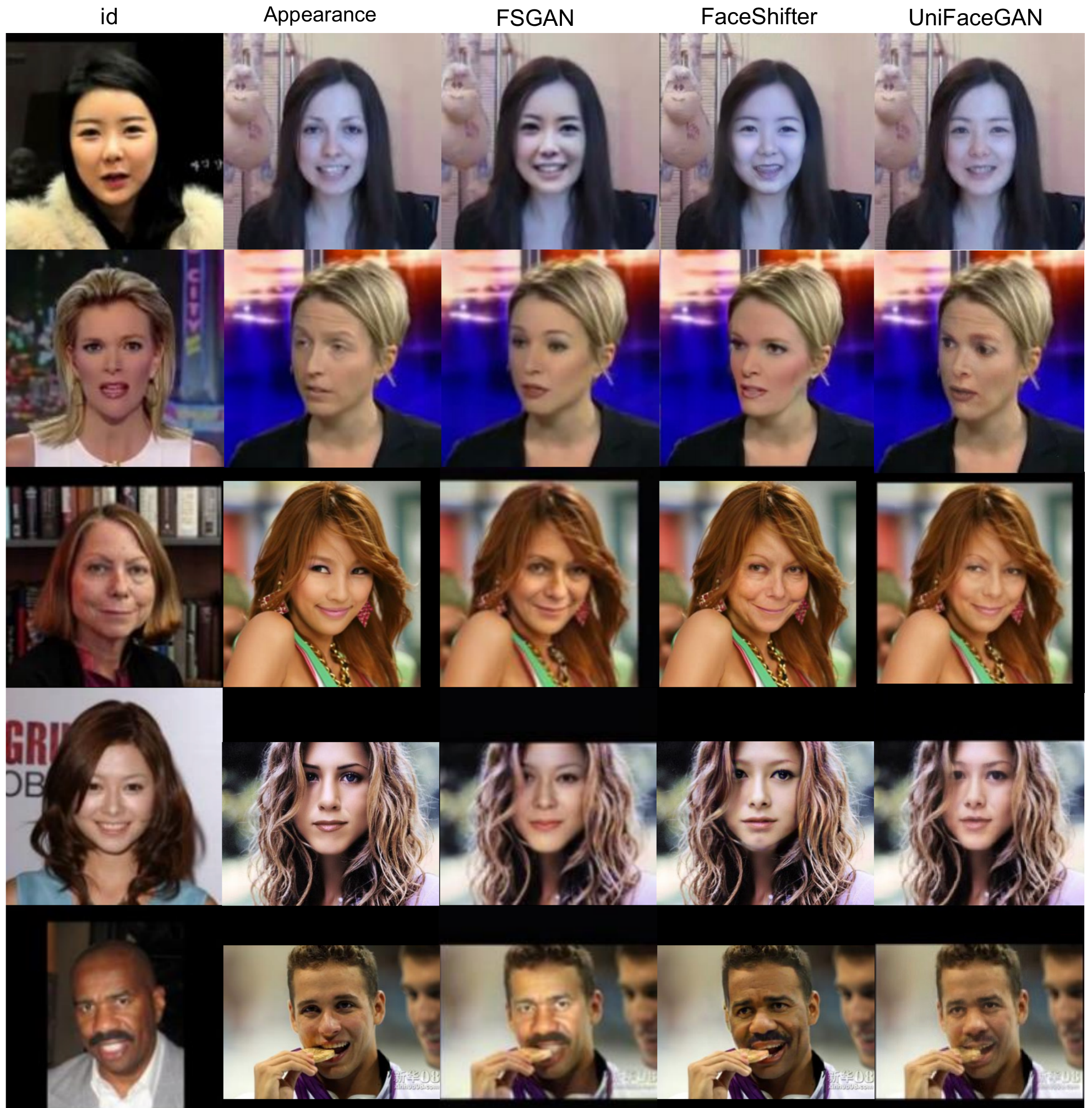}
	\vspace{-4mm}
	\caption{Comparisons with FaceShifter and FSGAN.}
	\label{fig:FaceShifter6Comp}
\end{figure}

%Firstly, we conduct quantitative evaluations on both face swapping and face reenactment to compare our method with the prior state-of-the-arts. Then we present the qualitative results of our new application `fully disentangled manipulation'.
\noindent\textbf{Face swapping.} When setting \(X_e^t = X_p^t\) and crop $M^t$ out of $X_p^t$, it falls into the face swapping problem. Fig.~\ref{fig:swapCompare} presents qualitative results against FaceShifter~\cite{li2019faceshifter} and FSGAN~\cite{nirkin2019fsgan}. UniFaceGAN generates high-fidelity results, even under challenging cases with varying ages, poses, gender and occlusions. The results of FSGAN are obviously blurred and over-smooth while FaceShiter suffers from striped flaws. Besides, quantitative results in Table~\ref{table:swapReenactCompare} also show that UniFaceGAN achieves better performance, which means our proposed UniFaceGAN is more photo-realistic and accurately captures the pose and expression information. To make fair comparisons, we compare our UniFaceGAN with FSGAN and Faceshifter using the cases presented in their paper and the results are shown in Figure.~\ref{fig:FaceShifter6Comp}. Note that we just infer on the images and do not fine tune our model on the corresponding datasets (\textit{e.g.} Caltech Occluded Faces in the Wild (COFW)~\cite{burgos2013robust}, IJB-C~\cite{maze2018iarpa}, VGGFace~\cite{parkhi2015deep} datasets, \textit{etc}). %Our UniFaceGAN achieves more photo-realistic 

\noindent\textbf{Face reenactment.}
To deal with the face reenactment task, we set \(X_e^t = X_p^t\) and crop $M^t$ out of $X_i^t$ or sample $X_i^t$ and $X_p^t$ in the same videos. Table~\ref{table:swapReenactCompare} and Fig~\ref{fig:reenactmentCompare} show the comparison results with state-of-the-art video face reenactment methods. FSGAN, FOMM~\cite{siarohin2019first} and Fast Bi-layer~\cite{zakharov2020fast} suffer from the blur background and UniFaceGAN settles this obstacle by explicitly providing an appearance hint. X2face shows a distorted face which is a common problem for warping-based methods. FTH can not animate the target person accurately, which demonstrates that our 3D-to-2D method is superior to the 2D landmark-based methods. 

\noindent\textbf{Fully disentangled manipulation.} When $X_i$, $X_p$ and $X_e$ are of different identities, we perform a novel fully disentangled manipulation task. The results in Fig.~\ref{fig:faceManipulation_new} show that the generated portrait image $Y$ successfully mixes information of identity, appearance and expression from different source images.

\begin{figure}[htb]
	\centering
	\includegraphics[width=\linewidth]{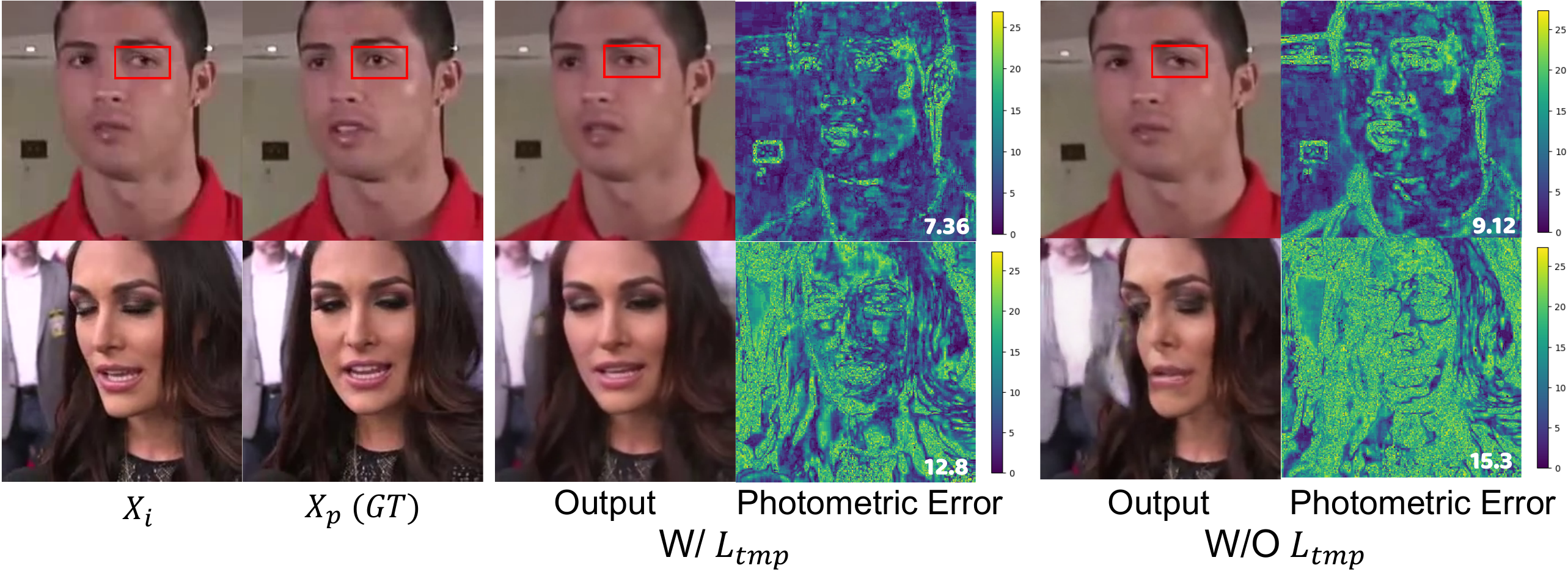}
	\vspace{-4mm}
	\caption{\small{Influence of $\mathcal{L}_{tmp}$ on Face Reenactment. $X_p$ is the ground truth because of the self-supervision scenario. The mean photometric error is listed bottom-right.}}
	\label{fig:errormap}
\end{figure}

\begin{figure*}[htb]
	\centering
	\subfigure[]{
		\begin{minipage}[b]{0.12\linewidth}
			\includegraphics[width=1.2\linewidth,height=1.2\linewidth]{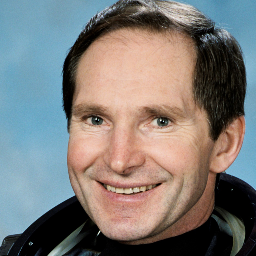}
		\end{minipage}
	}
	\subfigure[]{
		\begin{minipage}[b]{0.12\linewidth}
			\includegraphics[width=1.2\linewidth,height=1.2\linewidth]{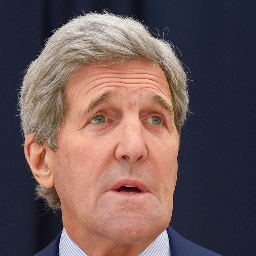}
		\end{minipage}
	}
	\subfigure[]{
		\begin{minipage}[b]{0.12\linewidth}
			\includegraphics[width=1.2\linewidth,height=1.2\linewidth]{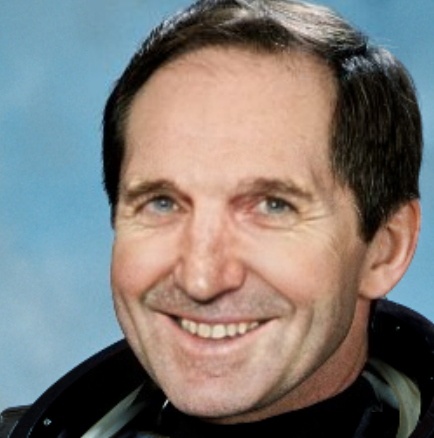}
		\end{minipage}
	}
	\subfigure[]{
		\begin{minipage}[b]{0.12\linewidth}
			\includegraphics[width=1.2\linewidth,height=1.2\linewidth]{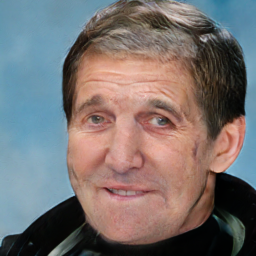}
		\end{minipage}
	}
	\subfigure[]{
		\begin{minipage}[b]{0.12\linewidth}
			\includegraphics[width=1.2\linewidth,height=1.2\linewidth]{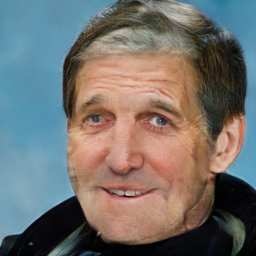}
		\end{minipage}
	}
	\subfigure[]{
		\begin{minipage}[b]{0.12\linewidth}
			\includegraphics[width=1.2\linewidth,height=1.2\linewidth]{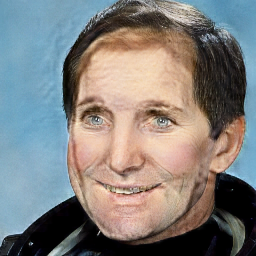} %W/O cross
		\end{minipage}
	}
	\subfigure[]{
		\begin{minipage}[b]{0.12\linewidth}
			\includegraphics[width=1.2\linewidth,height=1.2\linewidth]{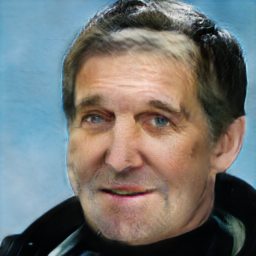} %W/O same
		\end{minipage}
	}
	\vspace{-3mm}
	\caption{Influence of RCN on Face Swapping. (a) Input appearance images. (b) Input identity images. Results: (c) RCN. (d)  AdaIN. (e) SPADE. (f) W/O cross-regional transfer term. (g) W/O same-regional retention term.}
	\label{fig:BSN_ablation}
\end{figure*}

%\begin{figure}[htp]
%	\centering
%	\includegraphics[width=\linewidth]{TFVGANSupply/longtermFlow.pdf}
%	\caption{Verification of the correctness of 3D temporal loss. (a) Previous frames at time $t-50$. (b) Current frames at time $t$. (c) Warping the current frames back to previous ones using the computed optical flow maps. (d) Visualization of the computed optical flow maps.}
%	\label{fig:longtermFlow}
%\end{figure}
\setlength{\tabcolsep}{1.4pt}
% For swapping, we use TVL1 in OpenCV to estimate optical flow for calculating $E_{\text{tmp}}$.
\subsection{Ablation Studies}
\noindent\textbf{Ablation studies on DTSS:} \qquad

W and W/O dynamic training sample selection (DTSS). DTSS selects the identity image $X_i$, the appearance image $X_p$ and the expression image $X_e$ in two ways. We shift the intra-video sampling ratio $\sigma$ manually and the results are listed in Table~\ref{table:DTSSAblation}. Intuitively, $E_{\text{tmp}}$ decreases with the increase of $\sigma$, since our 3D temporal loss is only applied in the intra-video case where $X_i$, $X_p$ and $X_e$ are sampled from the same video. SSIM is evaluated in the self-driving scenario and when $\sigma$ increases, SSIM increases naturally. The other evaluation criteria generally achieve the best scores when setting $\sigma$ to 0.5 for both the face swapping and the face reenactment task. Specifically, when the ratio $\sigma$ is set to 0, the whole framework falls into a face swapping pipeline. Similarly, setting $\sigma$ to 1, UniFaceGAN can only handle the face reenactment task. The comparison results between $\sigma=0.5$ and $\sigma=0$ or $\sigma=1$ indicate that this multi-task training is more effective than training the single task alone.% and benefits both of them.
%DTSS takes both image and video datasets as input to achieve high-fidelity and temporally consistent output. To validate its effectiveness,  we keep the network and datasets unchanged and train face swapping and face reenactment alone. Specifically, for face swapping, $X_i$, $X_p$ and $X_e$ are randomly selected from the whole dataset; for face reenactment, image datasets are regarded as single-frame videos. Temporal error $E_{\text{tmp}}$~\cite{huang2017real} is used to evaluate temporal consistency. Table~\ref{table:jointTrain} shows the joint training mechanism benefits both tasks. Training alone causes low-quality images (higher FID) and temporal inconsistency (higher $E_\text{tmp}$).
%Besides, we shift the image dataset ratio $\sigma$ manually and the results are shown in Fig 7. Intuitively, $E_\text{tmp}$ increases as the image dataset ratio $\sigma$ increases.

\noindent\textbf{Ablation studies on the loss functions:} \qquad

1) W and W/O temporal consistency loss. The ablation study of our 3D temporal loss $\mathcal{L}_{tmp}$ is available in Table~\ref{table:temporalAblation}. Here we evaluate $\mathcal{L}_{tmp}$ in the scenario of self-driving reenactment since only in this case we have the ground truth. Several cases are shown in Fig~\ref{fig:errormap} and photometric error~\cite{kim2018deep} is also computed between output and ground truth. Fig~\ref{fig:errormap} shows that $\mathcal{L}_{tmp}$ plays a significantly role especially when $X_i$ and $X_p$ have large pose variations. 

%2) Long Term Temporal Loss Constraint. We investigate the influence of long term temporal loss. Specifically, we set the input frames to be $X_i^t, X_p^t, X_e^t $ and $X_i^{t-k}, X_p^{t-k}, X_e^{t-k}$, $k \geq 1,$ and remain the datasets and the pipeline unchanged. The temporal interval $k$ is set to 1, 2, 4, 6, 8, 10, respectively, and the results are shown in Table~\ref{table:temporalAblation}. In general, as the temporal interval $k$ increases, $E_{\text{tmp1}}$ and $E_{\text{tmp2}}$ increase slightly. This proves the effectiveness of the proposed 3D temporal loss under the constraint of the long-term interval. For face swapping, $E_{\text{tmp1}}$ is computed using the optical flow map extracted by the TVL1 algorithm in OpenCV since this task has no ground truth. For face reenactment, $E_{\text{tmp2}}$ is still computed in the self-supervised scenario.

2) Dense flow vs. Sparse flow. vs. FlowNet2. To demonstrate the necessity of obtaining the dense optical flow maps using the barycentric coordinate interpolation, we compare UniFaceGAN with the models which only use sparse optical flow as temporal constraints. $\mathcal{L}_{tmp}$ is still calculated by Equation~\ref{con:calculateTMP}, but using the sparse optical flow map instead. Specifically, for each vertex of 3DMM reconstruction output, we project its optical flow value to the 2D plane orthogonally and predict the visibility according to the Z-Buffer value. Table~\ref{table:temporalAblation} gives the results of using sparse flow maps and the results show training with dense flow maps outperforms the sparse ones. In Table~\ref{table:temporalAblation}, we also compute the temporal loss using the optical flow estimation network FlowNet2~\cite{ilg2017flownet}. It shows that the optical flow computed by FlowNet2 leads to even worse results and this may be caused by the inaccurate facial optical flow estimation by FlowNet2.

\begin{table}[t]
	\begin{center}
		\caption{\small{Ablation studies for the 3D temporal loss.}} %W/O denotes training without the temporal loss. ``Dense" means the model trained with our interpolated optical flow map while ``Sparse" means only using the optical flow of the 3D vertex as the temporal supervision. ``FlowNet2" means using FlowNet2 to estimate the optical flow.}}
		\vspace{-2mm}
		\label{table:temporalAblation}
		%\begin{tabular}{llllllll}
		\resizebox{0.9\linewidth}{!}{
			\begin{tabular}{ccccccccc}
				%\hline
				\noalign{\smallskip}
				%\toprule[1pt]
				\toprule
				Task & Model    & FID\(\downarrow\) & SSIM\(\uparrow\) & $E_{\text{id}}$\(\downarrow\) & $E_{\text{pose}}$\(\downarrow\)  & $E_{\text{exp}}$\(\downarrow\) & $E_{\text{tmp}}$\(\downarrow\)\\ %& $E_{\text{tmp}}$\(\downarrow\) \\
				\noalign{\smallskip}
				%\hline
				%\toprule[1pt]
				\midrule
				\noalign{\smallskip}
				\multirow{4}*{Swapping}  & W/O  & 43.2 & - & 0.93 & \scriptsize(1.61,1.42,0.66) & 3.36 & 5.67\\
				~  &   FlowNet2 & 50.2 & -   & 1.34 & \scriptsize(1.63,1.73,0.66) & 3.76 & 5.94 \\
				~  &   Sparse & 38.2 & -   & 0.73 & \scriptsize(1.34,1.54,0.66) & 2.86 & 4.23 \\
				~  &   Dense(Ours)  & \textbf{34.1} & -   & \textbf{0.26} & \scriptsize(\textbf{1.12},\textbf{1.31},\textbf{0.65}) & \textbf{2.31} & \textbf{3.54} \\
				%\hline
				\midrule
				\multirow{4}*{Reenactment} & W/O   & 38.3 & 0.65 & 0.27 & \scriptsize(0.43,0.38,0.12)  & 3.93 & 6.91 \\
				~ &  FlowNet2 & 44.1 & 0.49 & 0.33 & \scriptsize(0.50,0.53,0.17) & 5.65 & 8.24 \\
				~ & Sparse & 30.0 & 0.75 & 0.17 & \scriptsize(0.46,0.39,0.14) & 3.52 & 6.54 \\
				~ & Dense(Ours)  & \textbf{24.4} & \textbf{0.82} & \textbf{0.12} & \scriptsize(\textbf{0.21},\textbf{0.25},\textbf{0.12}) & \textbf{3.12} & \textbf{3.65} \\
				%\toprule[1pt]
				\bottomrule
			\end{tabular}
		}
	\end{center}
\end{table}
%\vspace{-3mm}

\begin{figure*}[t] % select 2
	\centering
	\subfigure[]{
		\begin{minipage}[b]{0.12\linewidth}
			\includegraphics[width=1.2\linewidth,height=1.2\linewidth]{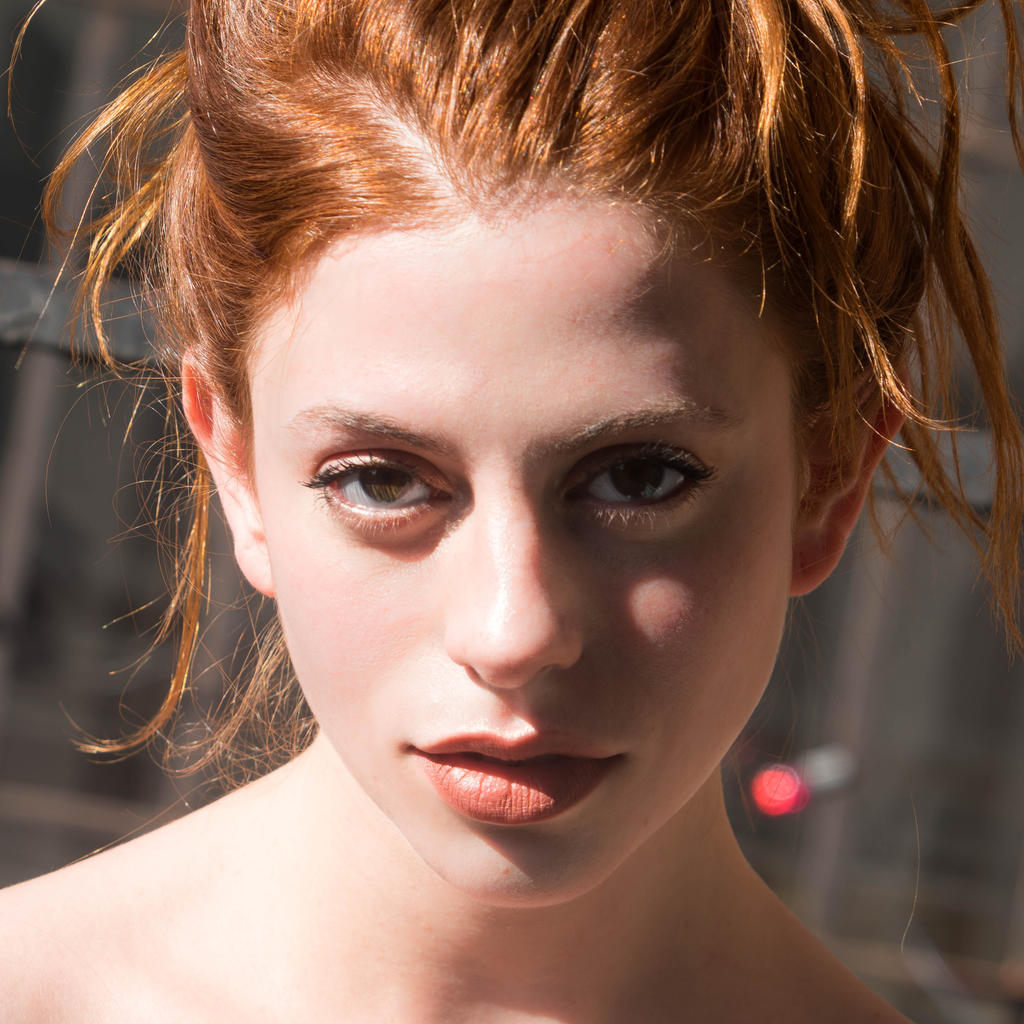}
		\end{minipage}
	}
	\subfigure[]{
		\begin{minipage}[b]{0.12\linewidth}
			\includegraphics[width=1.2\linewidth,height=1.2\linewidth]{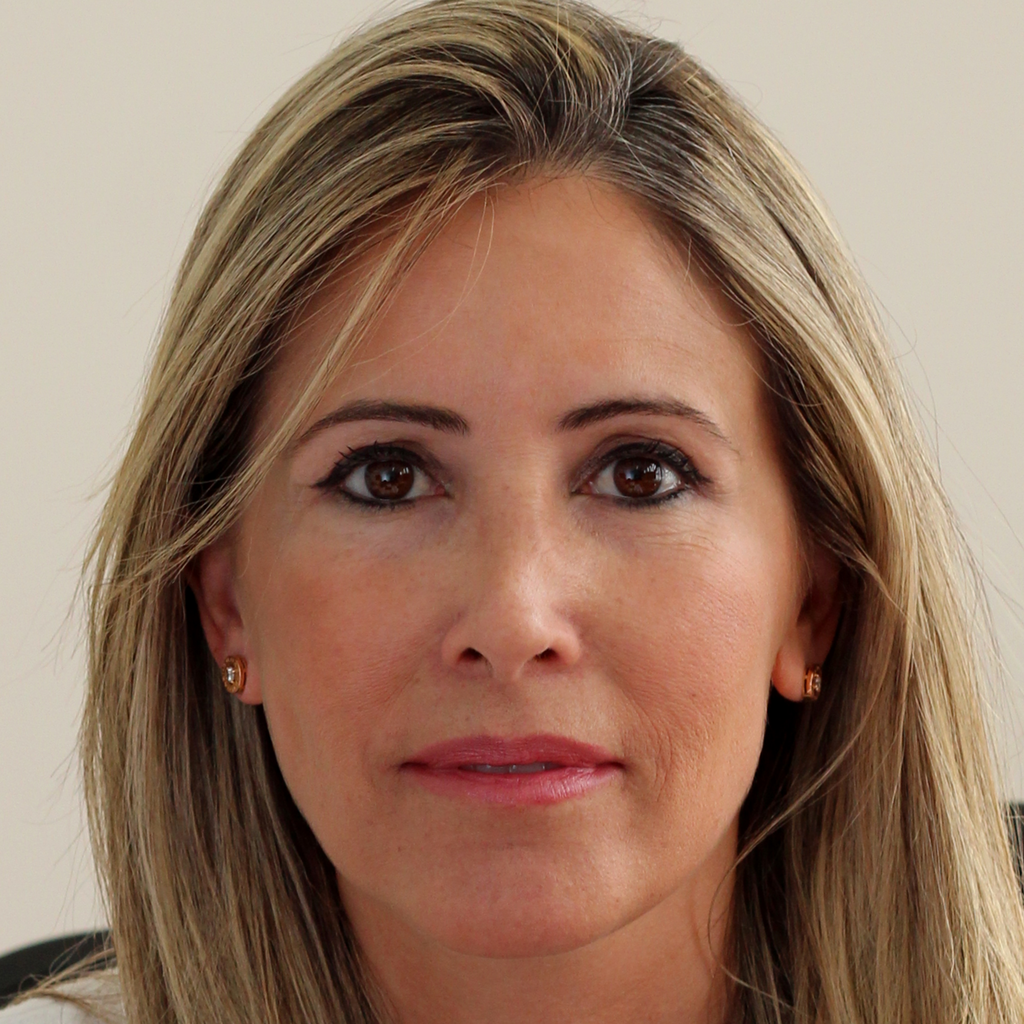}
		\end{minipage}
	}
	\subfigure[]{
		\begin{minipage}[b]{0.12\linewidth}
			\includegraphics[width=1.2\linewidth,height=1.2\linewidth]{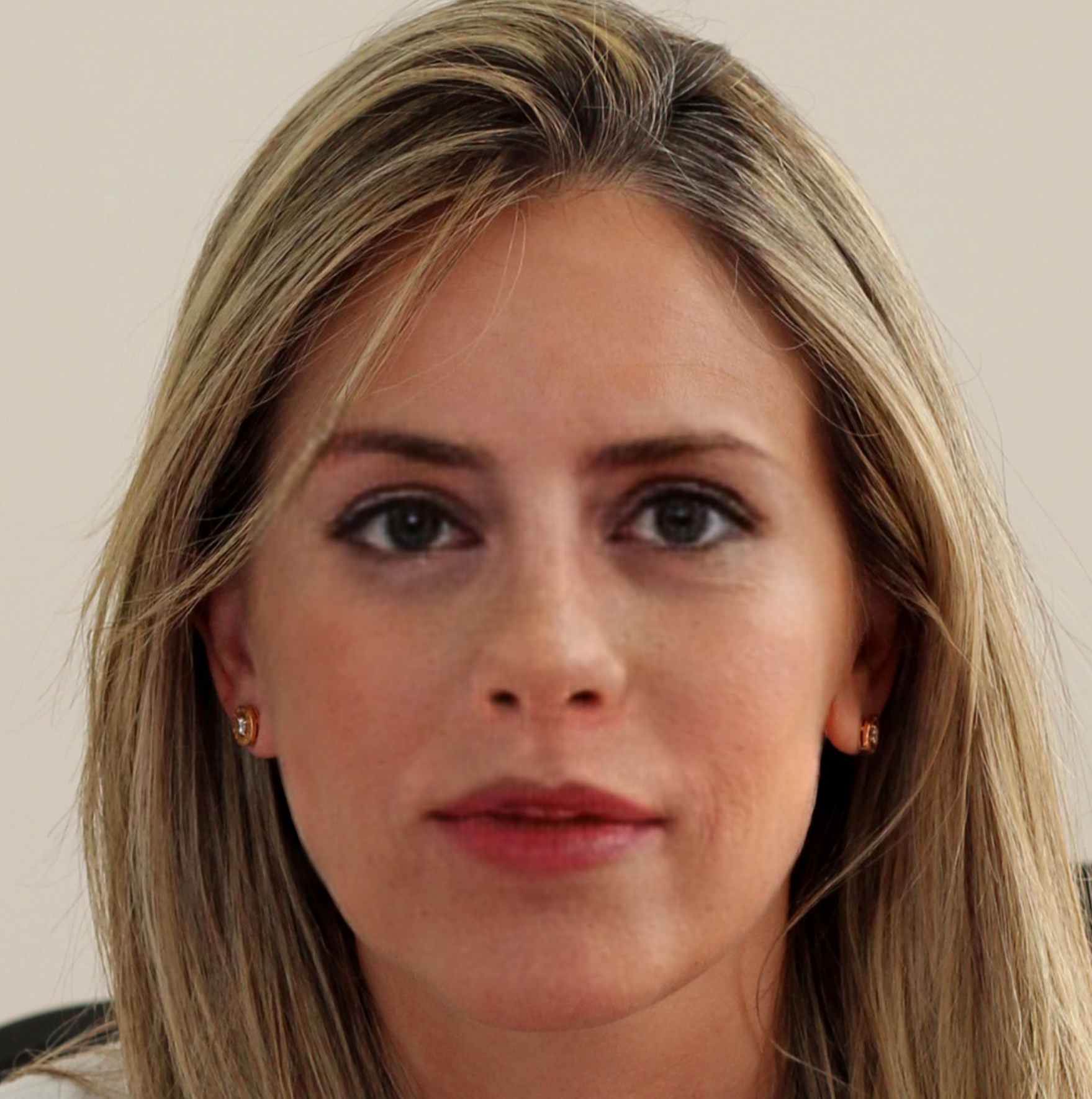}
		\end{minipage}
	} 
	\subfigure[]{
		\begin{minipage}[b]{0.12\linewidth}
			\includegraphics[width=1.2\linewidth,height=1.2\linewidth]{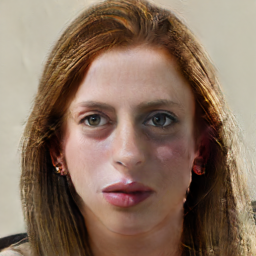}
		\end{minipage}
	} 
	\subfigure[]{
		\begin{minipage}[b]{0.12\linewidth}
			\includegraphics[width=1.2\linewidth,height=1.2\linewidth]{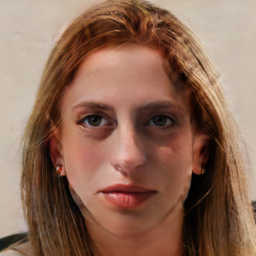}
		\end{minipage}
	}
	%\subfigure[]{
	%	\begin{minipage}[b]{0.12\linewidth}
	%		\includegraphics[width=1.2\linewidth,height=1.2\linewidth]{ablation/Loss/00041_00059_ep50.png}
	%	\end{minipage}
	%}
	\subfigure[]{
	\begin{minipage}[b]{0.12\linewidth}
		\includegraphics[width=1.2\linewidth,height=1.2\linewidth]{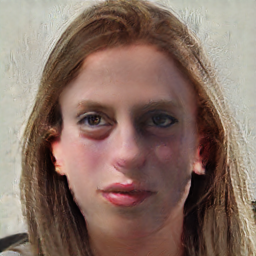}
	\end{minipage}
}
    \vspace{-3mm}
	\caption{\small{Ablation experiments for Appearance Preserving Loss $\mathcal{L}_{{app}}$ and Reconstruction Loss $\mathcal{L}_{{rec}}$. (a) Input identity images $X_i$. (b) Input appearance images $X_p$. (c) Results of UniFaceGAN (Ours). (d) Results trained W/O $\mathcal{L}_{{app}}$. (e) Results trained W/O $\mathcal{L}_{{rec}}$. (f) Results trained W/O self-attention.}}
	\label{fig:lossAbaFig}
\end{figure*}

3) W and W/O Appearance Preserving Loss or Reconstruction Loss. The experimental results in Table~\ref{table:lossAblation} show that either dropping $\mathcal{L}_{{app}}$ or $\mathcal{L}_{{rec}}$ leads to consistent performance degradation. The qualitative comparison results are also shown in Fig.~\ref{fig:lossAbaFig}. When dropping $\mathcal{L}_{{app}}$, the output images suffer from the blurry background (especially the hair regions); When dropping $\mathcal{L}_{{rec}}$, the quality of the output image is greatly affected (e.g., the unnatural transitions between face and background).

\begin{table}[]
	\centering
	\caption{\small{Ablation studies for $\mathcal{L}_{{app}}$ and $\mathcal{L}_{{rec}}$.}}
	\vspace{-2mm}
	\label{table:lossAblation}
	\resizebox{0.9\linewidth}{!}{
		\begin{tabular}{cccccccc}
			%\hline
			%\toprule[1pt]
			\toprule
			\multirow{2}{*}{Task}        & \multirow{2}{*}{$\mathcal{L}_{{app}}$} & \multirow{2}{*}{$\mathcal{L}_{{rec}}$} & \multirow{2}{*}{FID$\downarrow$} & \multirow{2}{*}{SSIM$\uparrow$} & \multirow{2}{*}{$E_{\text{id}}$$\downarrow$} & \multirow{2}{*}{$E_{\text{pose}}$$\downarrow$} & \multirow{2}{*}{$E_{\text{exp}}$$\downarrow$} \\
			&                     &                     &                      &                       &                      &                        &                       \\ 
			%\hline
			%\toprule[1pt]
			\midrule
			\multirow{3}{*}{swapping}    & \checkmark                   &                 &      41.4                &       -                &        0.39              &        \scriptsize (1.17,1.45,0.67)         &       2.71                \\
			&                    & \checkmark   &     38.7          &  -  &             0.34    &   \scriptsize (1.17,1.35,0.66)                                        &  2.54  \\
			& \checkmark  & \checkmark  & \textbf{34.1}  & -   & \textbf{0.26}   &  \scriptsize (\textbf{1.12},\textbf{1.31},\textbf{0.65})   & \textbf{2.31}  \\ 
			%\hline
			\midrule
			\multirow{3}{*}{Reenactment} & \checkmark &  &   30.2   &   0.71     &  0.19     &   \scriptsize(0.22,0.26,0.12)   &  4.98                             \\
			&                    & \checkmark                         &  28.6   &  0.74     &  0.15    &  \scriptsize(0.23,0.26,0.12)      & 4.03                      \\
			& \checkmark  & \checkmark          			  &  \textbf{24.4}    & \textbf{0.82} & \textbf{0.12} & \scriptsize(\textbf{0.21},\textbf{0.25},\textbf{0.12})   & \textbf{3.12}               \\ 
			%\hline
			%\toprule[1pt]
			\bottomrule
	\end{tabular}}
\end{table}

4) Removing the self-attention module in G. In the experiment, we add the self-attention module at the $64 \times 64$ resolution in the up-sampling part of the generator. To prove its necessity, we conduct the ablation studies without the self-attention module. The quantitative comparison results are shown Table~\ref{table:selfAttenAblation}. The results show that the inserted self-attention module improves the quality of the outputs greatly. For better illustrations, the qualitative comparison results are shown between Fig.~\ref{fig:lossAbaFig}(c) and Fig.~\ref{fig:lossAbaFig}(f).

5) Long Term Temporal Loss Constraint. We investigate the influence of long term temporal loss. Specifically, we set the input frames to be $X_i^t, X_p^t, X_e^t $ and $X_i^{t-k}, X_p^{t-k}, X_e^{t-k}$, $k \geq 1,$ and remain the datasets and the pipeline unchanged. The quantitative results are available in the supplementary material.

\begin{table}[]
	\centering
	\caption{\small{Ablation studies for the self-attention terms.}}
	\vspace{-2mm}
	\label{table:selfAttenAblation}
	\resizebox{\linewidth}{!}{
		\begin{tabular}{ccccccc}
			%\hline
			%\toprule[1pt]
			\toprule
			\multirow{2}{*}{Task}         & \multirow{2}{*}{self-attention} & \multirow{2}{*}{FID$\downarrow$} & \multirow{2}{*}{SSIM$\uparrow$} & \multirow{2}{*}{$E_{\text{id}}\downarrow$} & \multirow{2}{*}{$E_{\text{pose}}\downarrow$} & \multirow{2}{*}{$E_{\text{exp}}\downarrow$} \\
			&                                       &                      &                       &                      &                        &                       \\ 
			%\hline
			%\toprule[1pt]
			\midrule
			\multirow{2}{*}{swapping}                      &                 &      39.2               &       -                &        0.41              &        \scriptsize (1.18,1.51,0.67)         &       2.67                \\
			&          \checkmark           &     \textbf{34.1}       &  -  &     \textbf{0.26}   &   \scriptsize (\textbf{1.12},\textbf{1.31},\textbf{0.65})  &  \textbf{2.31}  \\
			\hline
			\multirow{2}{*}{Reenactment} &  &     29.7   &   0.64     &  0.23     &   \scriptsize(0.24,0.28,0.13)   &  3.65                             \\
			&  \checkmark   &  \textbf{24.4}   &  \textbf{0.82}  &  \textbf{0.12} &  \scriptsize(\textbf{0.21},\textbf{0.25},\textbf{0.12})   & \textbf{3.12}              \\
			%\hline
			%\toprule[1pt]
			\bottomrule
		\end{tabular}
	}
\end{table}

\begin{figure}[htpt] % select 2
	\centering
	%\subfigure[$X_i$]{
	\subfigure[]{
		\begin{minipage}[b]{0.16\linewidth}
			\includegraphics[width=1.2\linewidth,height=1.2\linewidth]{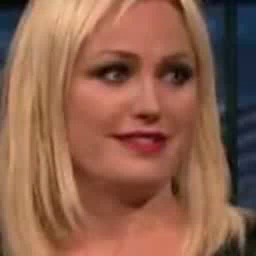}
		\end{minipage}
	}
	%\subfigure[$X_p$]{
	\subfigure[]{
		\begin{minipage}[b]{0.16\linewidth}
			\includegraphics[width=1.2\linewidth,height=1.2\linewidth]{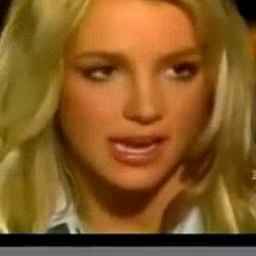}
		\end{minipage}
	}
	%\subfigure[\scriptsize{Full (Ours)}]{
	\subfigure[]{
		\begin{minipage}[b]{0.16\linewidth}
			\includegraphics[width=1.2\linewidth,height=1.2\linewidth]{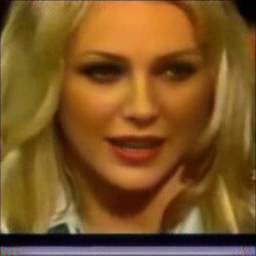}
		\end{minipage}
	} 
	%\subfigure[W/O cross]{
	%	\begin{minipage}[b]{0.12\linewidth}
		%	\includegraphics[width=1.2\linewidth,height=1.2\linewidth]{ablation/BSN/abRCNfaceNoface/noTransfer.png}
	%	\end{minipage}
	%} 
	%\subfigure[W/O same]{
%		\begin{minipage}[b]{0.12\linewidth}
	%		\includegraphics[width=1.2\linewidth,height=1.2\linewidth]{ablation/BSN/abRCNfaceNoface/noCopy.png}
	%	\end{minipage}
	%}
	%\subfigure[\scriptsize{Facial Only}]{
	\subfigure[]{
		\begin{minipage}[b]{0.16\linewidth}
			\includegraphics[width=1.2\linewidth,height=1.2\linewidth]{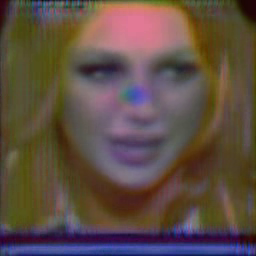}
		\end{minipage}
	}
	%\subfigure[\scriptsize{Non-facial Only}]{
	\subfigure[]{
		\begin{minipage}[b]{0.16\linewidth}
			\includegraphics[width=1.2\linewidth,height=1.2\linewidth]{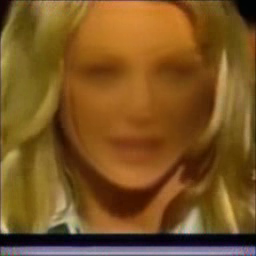}
		\end{minipage}
	}
	\vspace{-3mm}
	%\caption{\textcolor{red}{Ablation studies of RCN on Face Swapping. (a) The input identity image. (b) The input appearance image. Results: (c) Full RCN. (d)  W/O cross regional transfer term. (e) W/O same regional retention term. (f) Only facial terms. (g) Only non-facial terms items.}}
	\caption{\small{Ablation studies of RCN on Face Swapping. (a) The input identity image. (b) The input appearance image. Results: (c) Full RCN. (d) Only facial terms. (e) Only non-facial terms items.}}
	\label{fig:abRCNfaceNoface}
\end{figure}

\noindent\textbf{Ablation studies for RCN:} \qquad

1) RCN vs. AdaIN vs. SPADE. When replacing RCN with AdaIN or SPADE, the fidelity restoration of UniFaceGAN degrades substantially. The quantitative and qualitative comparison results in Table~\ref{table:rcnPart} (a) and Fig~\ref{fig:BSN_ablation} show that RCN leads to a more smooth transition between the facial and non-facial areas and looks more context-harmonious.

2) Are the cross-regional term and same-regional term necessary? We have added more ablation studies for RCN, \textit{i.e.,} experiments with only same-regional retention term or cross-regional transfer term. The quantitative and qualitative results are shown Table~\ref{table:rcnPart} (b) and Fig.~\ref{fig:BSN_ablation}. The experimental results demonstrate that both the same-regional retention term or cross-regional transfer terms are crucial to the photo-realistic outputs. The output without the cross-regional transfer term (Fig.~\ref{fig:BSN_ablation}(f))  has unsmooth boundary transitions between facial and non-facial parts. Besides, removing the same-regional retention term leads to the color contaminate, \textit{e.g.,} Fig.~\ref{fig:BSN_ablation}(g) facial part has the black shadow which may be transferred from the background of appearance image in Fig.~\ref{fig:BSN_ablation}(b).

3) Are just facial or non-facial related terms necessary? In Eqn.~\ref*{eqn:RCN}, the second ($\mathcal{T}(T_{l} \overline{H_{l}}, U_{l} H_{l})$) and the third ($U_l H_l$) terms contribute to the generated facial part and they are called facial related terms here. Similarly, the first ($\mathcal{T}(U_{l} H_{l}, T_{l} \bar{H}_{l})$) and the fourth ($T_l \overline {H}_l$) term are called non-facial related terms. We conduct comparison experiments with only facial region term or no-facial region transfer term. The quantitative and qualitative results are shown Table~\ref{table:rcnPart} (c) and Fig.~\ref{fig:abRCNfaceNoface}. The experimental results show that with only facial related items, the generated results suffer from the blurry background and color inconsistency and with only non-facial related items, the output can not generate the clear facial parts. All the loss term ablation studies are trained from scratch and remain the experimental settings unchanged.

\begin{table}[t]
	\centering
	\caption{\small{Ablation studies for RCN. (a) RCN \textit{v.s.} AdaIN \textit{v.s.} SPADE. (b) Are the cross-regional term and same-regional term necessary? (c) Are just facial or non-facial related terms necessary?}}
	\vspace{-2mm}
	\label{table:rcnPart}
	\resizebox{\linewidth}{!}{
		\begin{tabular}{cccccccc}
			%\hline
			%\toprule[1pt]
			\toprule
			 (a) & Mode   & FID\(\downarrow\) & SSIM\(\uparrow\) & $E_{\text{id}}$\(\downarrow\) & $E_{\text{pose}}$\(\downarrow\)  & $E_{\text{exp}}$\(\downarrow\) \\ %& $E_{\text{tmp}}$\(\downarrow\) \\
			\noalign{\smallskip}
			%\hline
			%\toprule[1pt]
			\midrule
			\noalign{\smallskip}
			\multirow{3}*{Swapping}  & RCN(Ours)   & \textbf{34.1} & - & \textbf{0.26} & \scriptsize(1.12,\textbf{1.31},\textbf{0.65}) & \textbf{2.31} \\
			~  &    AdaIN  & 47.2 & -   & 0.59 & \scriptsize(1.12,1.43,0.66) & 2.72 \\
			~  &   SPADE  & 39.5 & -   & 0.50 & \scriptsize(\textbf{1.11},1.33,0.65) & 2.55 \\
			%\hline
			\midrule
			\multirow{3}*{Reenactment} & RCN(Ours)    & \textbf{24.4} & \textbf{0.82} & \textbf{0.12} & \scriptsize(\textbf{0.21},\textbf{0.25},\textbf{0.12})  & \textbf{3.12} \\
			~ &  AdaIN  & 36.9 & 0.72 & 0.18 & \scriptsize(0.21,0.32,0.12) & 3.52 \\
			~ &  SPADE  & 32.1 & 0.76 & 0.18 & \scriptsize(0.21,0.27,0.12) & 3.38 \\
			%\toprule[1pt]
			%\bottomrule
			\midrule
			%\toprule
			\midrule
			(b)        & Mode & FID$\downarrow$ & SSIM$\uparrow$ & $E_{\text{id}}\downarrow$ &  $E_{\text{pose}}\downarrow$ & $E_{\text{exp}}\downarrow$ \\
			%\multirow{2}{*}{Task}        & \multirow{2}{*}{$\mathcal{L}_{{cross}}$} & \multirow{2}{*}{FID$\downarrow$} & \multirow{2}{*}{SSIM$\uparrow$} & \multirow{2}{*}{$E_{\text{id}}\downarrow$} & \multirow{2}{*}{$E_{\text{pose}}\downarrow$} & \multirow{2}{*}{$E_{\text{exp}}\downarrow$} \\
			%&                     &                     &                      &                       &                      &                        &                       \\ 
			%\hline
			%\toprule[1pt]
			\midrule
			\multirow{3}{*}{swapping}    & Cross only                              &      44.2         &       -                &        0.51              &        \scriptsize (1.20,1.47,0.67)         &       3.06                \\
			&      Same only                 &     40.3          &  -  &          0.45    &   \scriptsize (1.18,1.41,0.66)                             &  2.63  \\
			& Full(Ours)    & \textbf{34.1}  & -   & \textbf{0.26}   &  \scriptsize (\textbf{1.12},\textbf{1.31},\textbf{0.65})   & \textbf{2.31}  \\ \hline
			\multirow{3}{*}{Reenactment} & Cross only &     31.6   &   0.58     &  0.23     &   \scriptsize(0.24,0.26,0.13)   &  3.70                             \\
			&      Same only                      &  30.2   &  0.64     &  0.17    &  \scriptsize(0.25,0.26,0.12)      & 3.51                      \\
			& Full(Ours)			  &  \textbf{24.4}    & \textbf{0.82} & \textbf{0.12} & \scriptsize(\textbf{0.21},\textbf{0.25},\textbf{0.12})   & \textbf{3.12}               \\ 
			%\hline
			%\toprule[1pt]
			\midrule
			\midrule
		%	\multirow{2}{*}{Task}        & \multirow{2}{*}{$\mathcal{L}_{{facial}}$} & \multirow{2}{*}{FID$\downarrow$} & \multirow{2}{*}{SSIM$\uparrow$} & \multirow{2}{*}{$E_{\text{id}}\downarrow$} & \multirow{2}{*}{$E_{\text{pose}}\downarrow$} & \multirow{2}{*}{$E_{\text{exp}}\downarrow$} \\
			
			(c)        &  Mode & FID$\downarrow$ & SSIM$\uparrow$ & $E_{\text{id}}\downarrow$ &  $E_{\text{pose}}\downarrow$ & $E_{\text{exp}}\downarrow$ \\
			%&                     &                     &                      &                       &                      &                        &                       \\ 
			\midrule
			\multirow{3}{*}{swapping}    & Facial only                               &      50.9         &       -                &        0.73              &        \scriptsize (1.15,1.40,0.67)         &       4.23                \\
			&    Non-facial only                   &     49.3          &  -  &          0.86    &   \scriptsize (1.15,1.38,0.66)                             &  5.34  \\
			& Full(Ours)    & \textbf{34.1}  & -   & \textbf{0.26}   &  \scriptsize (\textbf{1.12},\textbf{1.31},\textbf{0.65})   & \textbf{2.31}  \\ \hline
			\multirow{3}{*}{Reenactment} & Facial only  &    37.3   &   0.29     &  0.40     &   \scriptsize(0.24,0.28,0.13)   &  4.32                             \\
			&   Non-facial only                                         &  35.1   &  0.36     &  0.35    &  \scriptsize(0.24,0.28,0.12)      & 3.83                      \\
			& Full(Ours)           			  &  \textbf{24.4}    & \textbf{0.82} & \textbf{0.12} & \scriptsize(\textbf{0.21},\textbf{0.25},\textbf{0.12})   & \textbf{3.12}               \\ 
			%\hline
			%\toprule[1pt]
			\bottomrule
	\end{tabular}}
\end{table}

%\noindent\textbf{Optical Flow Map Visualization.} %Fig.~\ref{fig:videoFlow} shows the optical flow maps between two consecutive frames using barycentric coordinate interpolation. In addition, 
%We compute the optical flow map between frames at time $t$ and time $t-50$ and use it to warp the frame at time $t$ back to time $t-50$. The results in Fig.~\ref{fig:longtermFlow} show that the computed  the optical flow map effectively warps frames at $t$ back to $t-50$, which verifies the correctness of our 3D temporal loss.

\section{Conclusions}
In this work, we have proposed a unified framework called UniFaceGAN for handling multiple video portrait manipulation tasks and generating temporally consistent outputs. To be specific, we carefully design a Dynamic Training Sample Selection mechanism to facilitate this multi-task training. A novel 3D temporal loss is proposed to enforce the visual consistency in synthesized videos. Besides, we further introduce a region-aware conditional normalization (RCN) layer to achieve the better face blending. The extensive experiments have demonstrated the consistent superiority of the proposed framework across multiple tasks over the state-of-the-art methods.

\ifCLASSOPTIONcaptionsoff
  \newpage
\fi



%\bibliographystyle{IEEEtran}
%\begin{thebibliography}{1}

%\small
%\bibliographystyle{unsrt} % For reference sorting
%\bibliography{refs.bib}
%\end{thebibliography}
%\clearpage
%\small
\scriptsize	
\bibliographystyle{unsrt} % For reference sorting
\bibliography{refs.bib}

\begin{thebibliography}{10}

\bibitem{kim2018deep}
Hyeongwoo Kim, Pablo Garrido, Ayush Tewari, Weipeng Xu, Justus Thies, Matthias
  Nie{\ss}ner, Patrick P{\'e}rez, Christian Richardt, Michael Zollh{\"o}fer,
  and Christian Theobalt.
\newblock Deep video portraits.
\newblock {\em ACM Transactions on Graphics (TOG)}, 37(4):1--14, 2018.

\bibitem{olszewski2017realistic}
Kyle Olszewski, Zimo Li, Chao Yang, Yi~Zhou, Ronald Yu, Zeng Huang, Sitao
  Xiang, Shunsuke Saito, Pushmeet Kohli, and Hao Li.
\newblock Realistic dynamic facial textures from a single image using gans.
\newblock In {\em Proceedings of the IEEE International Conference on Computer
  Vision}, pages 5429--5438, 2017.

\bibitem{korshunova2017fast}
Iryna Korshunova, Wenzhe Shi, Joni Dambre, and Lucas Theis.
\newblock Fast face-swap using convolutional neural networks.
\newblock In {\em Proceedings of the IEEE International Conference on Computer
  Vision}, pages 3677--3685, 2017.

\bibitem{nirkin2018face}
Yuval Nirkin, Iacopo Masi, Anh~Tran Tuan, Tal Hassner, and Gerard Medioni.
\newblock On face segmentation, face swapping, and face perception.
\newblock In {\em 2018 13th IEEE International Conference on Automatic Face \&
  Gesture Recognition (FG 2018)}, pages 98--105. IEEE, 2018.

\bibitem{jin2017cyclegan}
Xiaohan Jin, Ye~Qi, and Shangxuan Wu.
\newblock Cyclegan face-off.
\newblock {\em arXiv preprint arXiv:1712.03451}, 2017.

\bibitem{nirkin2019fsgan}
Yuval Nirkin, Yosi Keller, and Tal Hassner.
\newblock Fsgan: Subject agnostic face swapping and reenactment.
\newblock In {\em Proceedings of the IEEE International Conference on Computer
  Vision}, pages 7184--7193, 2019.

\bibitem{garrido2014automatic}
Pablo Garrido, Levi Valgaerts, Ole Rehmsen, Thorsten Thormahlen, Patrick Perez,
  and Christian Theobalt.
\newblock Automatic face reenactment.
\newblock In {\em Proceedings of the IEEE Conference on Computer Vision and
  Pattern Recognition}, pages 4217--4224, 2014.

\bibitem{thies2016face2face}
Justus Thies, Michael Zollhofer, Marc Stamminger, Christian Theobalt, and
  Matthias Nie{\ss}ner.
\newblock Face2face: Real-time face capture and reenactment of rgb videos.
\newblock In {\em Proceedings of the IEEE conference on computer vision and
  pattern recognition}, pages 2387--2395, 2016.

\bibitem{suwajanakorn2017synthesizing}
Supasorn Suwajanakorn, Steven~M Seitz, and Ira Kemelmacher-Shlizerman.
\newblock Synthesizing obama: learning lip sync from audio.
\newblock {\em ACM Transactions on Graphics (TOG)}, 36(4):1--13, 2017.

\bibitem{averbuch2017bringing}
Hadar Averbuch-Elor, Daniel Cohen-Or, Johannes Kopf, and Michael~F Cohen.
\newblock Bringing portraits to life.
\newblock {\em ACM Transactions on Graphics (TOG)}, 36(6):196, 2017.

\bibitem{wiles2018x2face}
Olivia Wiles, A~Sophia~Koepke, and Andrew Zisserman.
\newblock X2face: A network for controlling face generation using images,
  audio, and pose codes.
\newblock In {\em Proceedings of the European Conference on Computer Vision
  (ECCV)}, pages 670--686, 2018.

\bibitem{siarohin2019animating}
Aliaksandr Siarohin, St{\'e}phane Lathuili{\`e}re, Sergey Tulyakov, Elisa
  Ricci, and Nicu Sebe.
\newblock Animating arbitrary objects via deep motion transfer.
\newblock In {\em Proceedings of the IEEE Conference on Computer Vision and
  Pattern Recognition}, pages 2377--2386, 2019.

\bibitem{wu2018reenactgan}
Wayne Wu, Yunxuan Zhang, Cheng Li, Chen Qian, and Chen Change~Loy.
\newblock Reenactgan: Learning to reenact faces via boundary transfer.
\newblock In {\em Proceedings of the European Conference on Computer Vision
  (ECCV)}, pages 603--619, 2018.

\bibitem{zakharov2019few}
Egor Zakharov, Aliaksandra Shysheya, Egor Burkov, and Victor Lempitsky.
\newblock Few-shot adversarial learning of realistic neural talking head
  models.
\newblock In {\em Proceedings of the IEEE International Conference on Computer
  Vision}, pages 9459--9468, 2019.

\bibitem{zhao2017dual}
Jian Zhao, Lin Xiong, Panasonic~Karlekar Jayashree, Jianshu Li, Fang Zhao,
  Zhecan Wang, Panasonic~Sugiri Pranata, Panasonic~Shengmei Shen, Shuicheng
  Yan, and Jiashi Feng.
\newblock Dual-agent gans for photorealistic and identity preserving profile
  face synthesis.
\newblock In {\em Advances in neural information processing systems}, pages
  66--76, 2017.

\bibitem{blanz1999morphable}
Volker Blanz and Thomas Vetter.
\newblock A morphable model for the synthesis of 3d faces.
\newblock In {\em Proceedings of the 26th annual conference on Computer
  graphics and interactive techniques}, pages 187--194, 1999.

\bibitem{zhu2016face}
Xiangyu Zhu, Zhen Lei, Xiaoming Liu, Hailin Shi, and Stan~Z Li.
\newblock Face alignment across large poses: A 3d solution.
\newblock In {\em Proceedings of the IEEE conference on computer vision and
  pattern recognition}, pages 146--155, 2016.

\bibitem{nagano2018pagan}
Koki Nagano, Jaewoo Seo, Jun Xing, Lingyu Wei, Zimo Li, Shunsuke Saito, Aviral
  Agarwal, Jens Fursund, and Hao Li.
\newblock pagan: real-time avatars using dynamic textures.
\newblock {\em ACM Transactions on Graphics (TOG)}, 37(6):1--12, 2018.

\bibitem{ha2019marionette}
Sungjoo Ha, Martin Kersner, Beomsu Kim, Seokjun Seo, and Dongyoung Kim.
\newblock Marionette: Few-shot face reenactment preserving identity of unseen
  targets.
\newblock {\em arXiv preprint arXiv:1911.08139}, 2019.

\bibitem{li2019faceshifter}
Lingzhi Li, Jianmin Bao, Hao Yang, Dong Chen, and Fang Wen.
\newblock Faceshifter: Towards high fidelity and occlusion aware face swapping.
\newblock {\em arXiv preprint arXiv:1912.13457}, 2019.

\bibitem{blanz2004exchanging}
Volker Blanz, Kristina Scherbaum, Thomas Vetter, and Hans-Peter Seidel.
\newblock Exchanging faces in images.
\newblock In {\em Computer Graphics Forum}, volume~23, pages 669--676. Wiley
  Online Library, 2004.

\bibitem{bao2018towards}
Jianmin Bao, Dong Chen, Fang Wen, Houqiang Li, and Gang Hua.
\newblock Towards open-set identity preserving face synthesis.
\newblock In {\em Proceedings of the IEEE Conference on Computer Vision and
  Pattern Recognition}, pages 6713--6722, 2018.

\bibitem{zhu2017face}
Xiangyu Zhu, Xiaoming Liu, Zhen Lei, and Stan~Z Li.
\newblock Face alignment in full pose range: A 3d total solution.
\newblock {\em IEEE transactions on pattern analysis and machine intelligence},
  41(1):78--92, 2017.

\bibitem{paysan20093d}
Pascal Paysan, Reinhard Knothe, Brian Amberg, Sami Romdhani, and Thomas Vetter.
\newblock A 3d face model for pose and illumination invariant face recognition.
\newblock In {\em 2009 Sixth IEEE International Conference on Advanced Video
  and Signal Based Surveillance}, pages 296--301. Ieee, 2009.

\bibitem{cao2013facewarehouse}
Chen Cao, Yanlin Weng, Shun Zhou, Yiying Tong, and Kun Zhou.
\newblock Facewarehouse: A 3d facial expression database for visual computing.
\newblock {\em IEEE Transactions on Visualization and Computer Graphics},
  20(3):413--425, 2013.

\bibitem{huang2017arbitrary}
Xun Huang and Serge Belongie.
\newblock Arbitrary style transfer in real-time with adaptive instance
  normalization.
\newblock In {\em Proceedings of the IEEE International Conference on Computer
  Vision}, pages 1501--1510, 2017.

\bibitem{park2019semantic}
Taesung Park, Ming-Yu Liu, Ting-Chun Wang, and Jun-Yan Zhu.
\newblock Semantic image synthesis with spatially-adaptive normalization.
\newblock In {\em Proceedings of the IEEE Conference on Computer Vision and
  Pattern Recognition}, pages 2337--2346, 2019.

\bibitem{wang2018high}
Ting-Chun Wang, Ming-Yu Liu, Jun-Yan Zhu, Andrew Tao, Jan Kautz, and Bryan
  Catanzaro.
\newblock High-resolution image synthesis and semantic manipulation with
  conditional gans.
\newblock In {\em Proceedings of the IEEE conference on computer vision and
  pattern recognition}, pages 8798--8807, 2018.

\bibitem{mao2017least}
Xudong Mao, Qing Li, Haoran Xie, Raymond~YK Lau, Zhen Wang, and Stephen
  Paul~Smolley.
\newblock Least squares generative adversarial networks.
\newblock In {\em Proceedings of the IEEE International Conference on Computer
  Vision}, pages 2794--2802, 2017.

\bibitem{lim2017geometric}
Jae~Hyun Lim and Jong~Chul Ye.
\newblock Geometric gan.
\newblock {\em arXiv preprint arXiv:1705.02894}, 2017.

\bibitem{dong2018supervision}
Xuanyi Dong, Shoou-I Yu, Xinshuo Weng, Shih-En Wei, Yi~Yang, and Yaser Sheikh.
\newblock Supervision-by-registration: An unsupervised approach to improve the
  precision of facial landmark detectors.
\newblock In {\em Proceedings of the IEEE Conference on Computer Vision and
  Pattern Recognition}, pages 360--368, 2018.

\bibitem{ilg2017flownet}
Eddy Ilg, Nikolaus Mayer, Tonmoy Saikia, Margret Keuper, Alexey Dosovitskiy,
  and Thomas Brox.
\newblock Flownet 2.0: Evolution of optical flow estimation with deep networks.
\newblock In {\em Proceedings of the IEEE conference on computer vision and
  pattern recognition}, pages 2462--2470, 2017.

\bibitem{chung2018voxceleb2}
Joon~Son Chung, Arsha Nagrani, and Andrew Zisserman.
\newblock Voxceleb2: Deep speaker recognition.
\newblock {\em arXiv preprint arXiv:1806.05622}, 2018.

\bibitem{heusel2017gans}
Martin Heusel, Hubert Ramsauer, Thomas Unterthiner, Bernhard Nessler, and Sepp
  Hochreiter.
\newblock Gans trained by a two time-scale update rule converge to a local nash
  equilibrium.
\newblock In {\em Advances in neural information processing systems}, pages
  6626--6637, 2017.

\bibitem{wang2004image}
Zhou Wang, Alan~C Bovik, Hamid~R Sheikh, and Eero~P Simoncelli.
\newblock Image quality assessment: from error visibility to structural
  similarity.
\newblock {\em IEEE transactions on image processing}, 13(4):600--612, 2004.

\bibitem{burgos2013robust}
Xavier~P Burgos-Artizzu, Pietro Perona, and Piotr Doll{\'a}r.
\newblock Robust face landmark estimation under occlusion.
\newblock In {\em Proceedings of the IEEE international conference on computer
  vision}, pages 1513--1520, 2013.

\bibitem{maze2018iarpa}
Brianna Maze, Jocelyn Adams, James~A Duncan, Nathan Kalka, Tim Miller, Charles
  Otto, Anil~K Jain, W~Tyler Niggel, Janet Anderson, Jordan Cheney, et~al.
\newblock Iarpa janus benchmark-c: Face dataset and protocol.
\newblock In {\em 2018 International Conference on Biometrics (ICB)}, pages
  158--165. IEEE, 2018.

\bibitem{parkhi2015deep}
Omkar~M Parkhi, Andrea Vedaldi, and Andrew Zisserman.
\newblock Deep face recognition.
\newblock 2015.

\bibitem{siarohin2019first}
Aliaksandr Siarohin, St{\'e}phane Lathuili{\`e}re, Sergey Tulyakov, Elisa
  Ricci, and Nicu Sebe.
\newblock First order motion model for image animation.
\newblock In {\em Advances in Neural Information Processing Systems}, pages
  7137--7147, 2019.

\bibitem{zakharov2020fast}
Egor Zakharov, Aleksei Ivakhnenko, Aliaksandra Shysheya, and Victor Lempitsky.
\newblock Fast bi-layer neural synthesis of one-shot realistic head avatars.
\newblock In {\em European Conference on Computer Vision}, pages 524--540.
  Springer, 2020.

\end{thebibliography}

\end{document}